\documentclass{article}

\usepackage[nonatbib,final]{neurips_2022}

\usepackage[utf8]{inputenc} 
\usepackage[T1]{fontenc}    
\usepackage{url}            
\usepackage{booktabs}       
\usepackage{amsfonts}       
\usepackage{nicefrac}       
\usepackage{microtype}      

\usepackage{graphicx}
\usepackage{cite}
\usepackage{color}
\usepackage{wrapfig}
\usepackage{epsfig}
\usepackage{graphicx}
\usepackage{amsmath}
\usepackage{amssymb}
\usepackage{multirow}
\usepackage{algorithm}
\usepackage{algorithmic}
\usepackage{arydshln}
\usepackage{threeparttable}
\usepackage{colortbl}
\usepackage{enumitem}
\usepackage{siunitx}
\usepackage{bm}
\usepackage{array}
\usepackage[table]{xcolor}
\usepackage{colortbl}
\usepackage{caption}
\usepackage{dblfloatfix}
\usepackage{amssymb}
\usepackage{pifont}
\usepackage{amssymb}
\usepackage{float}

\usepackage{listings}
\usepackage[export]{adjustbox}
\usepackage{makecell}
\usepackage{centernot}
\usepackage{subfig}
\usepackage{math_symbols}
\usepackage{bm}
\usepackage{enumitem}
\usepackage[accsupp]{axessibility}  

\usepackage{xr}
\usepackage{fancyhdr}
\usepackage[nopar]{lipsum}
\usepackage{gensymb}

\usepackage{subfiles}

\definecolor{mygray}{gray}{.9}
\definecolor{ggray}{RGB}{127,127,127}
\definecolor{reda}{RGB}{192,0,0}
\definecolor{redb}{RGB}{217,148,143}
\definecolor{myyellow}{RGB}{190,144,0}
\definecolor{mygreen}{RGB}{93,173,85}
\definecolor{myblue}{RGB}{30,90,100}
\definecolor{demphcolor}{RGB}{100,100,100}
\newcommand{\best}[1]{{\textcolor{red}{\textbf{#1}}}}
\newcommand{\second}[1]{{\textcolor{blue}{#1}}}
\newcommand{\tabincell}[2]{\begin{tabular}{@{}#1@{}}#2\end{tabular}}

\newcommand{\tablestyle}[2]{\setlength{\tabcolsep}{#1}\renewcommand{\arraystretch}{#2}\centering\footnotesize}

\newcolumntype{x}[1]{>{\centering\arraybackslash}p{#1pt}}
\newcolumntype{I}{!{\vrule width 1pt}}
\newcolumntype{d}[1]{>{\raggedright\arraybackslash}p{#1pt}}
\newcolumntype{b}[1]{>{\raggedleft\arraybackslash}p{#1pt}}

\makeatletter
\newcommand\footnoteref[1]{\protected@xdef\@thefnmark{\ref{#1}}\@footnotemark}

\makeatother

\usepackage{xspace}
\makeatletter
\DeclareRobustCommand\onedot{\futurelet\@let@token\@onedot}
\def\@onedot{\ifx\@let@token.\else.\null\fi\xspace}

\def\eg{\textit{e.g}\onedot} 
\def\ie{\textit{i.e}\onedot} 
 
\def\etc{\textit{etc}\onedot} 
\def\wrt{w.r.t\onedot} 
\def\etal{\textit{et al}\onedot}

\makeatother

\makeatletter
\newcommand{\thickhline}{%
    \noalign {\ifnum 0=`}\fi \hrule height 1pt
    \futurelet \reserved@a \@xhline
}



\usepackage[pagebackref=true,breaklinks=true,letterpaper=true,colorlinks,bookmarks=false]{hyperref}

\title{Towards Versatile Embodied Navigation}

\author{%
  Hanqing Wang\textsuperscript{\textnormal{1, 2}}\qquad
  Wei Liang\textsuperscript{\textnormal{1,4}$^*$}\qquad
  Luc Van Gool\textsuperscript{\textnormal{2}}\qquad
  Wenguan Wang\textsuperscript{\textnormal{3}\thanks{Corresponding authors.}}\\
  \\
  \textsuperscript{\textnormal{1}}Beijing Institute of Technology\ \ \textsuperscript{\textnormal{2}}Computer Vision Lab, ETH Zurich \\ \textsuperscript{\textnormal{3}}ReLER, AAII, University of Technology Sydney\\
  \textsuperscript{\textnormal{4}}Yangtze Delta Region Academy of Beijing Institute of Technology, Jiaxing\\
  \\
  Project page: \url{https://github.com/hanqingwangai/VXN}\\
}

\begin{document}

\maketitle

\begin{abstract}
With the emergence of varied visual navigation tasks (\eg, image-/object-/audio-goal and vision-language navigation) that specify the target in different ways, the community has made appealing advances in training specialized agents capable of handling individual navigation tasks well.
Given plenty of embodied navigation tasks and task-specific solutions, we address a more fundamental question: \textit{can we learn a single powerful agent that masters not one but multiple navigation~tasks concurrently}? First, we propose \textbf{\texttt{VXN}}, a large-scale 3D dataset that instantiates~four classic navigation tasks in standardized, continuous, and audiovisual-rich environ- ments. Second, we propose$_{\!}$ \textsc{Vienna}, a \underline{v}ersat\underline{i}le \underline{e}mbodied \underline{n}avigatio\underline{n} \underline{a}gent that simultaneously learns to perform the four navigation tasks with one model.$_{\!}$ Building$_{\!}$ upon$_{\!}$ a full-attentive architecture, \textsc{Vienna} formulates various navigation tasks as a unified, \textit{parse-and-query} procedure: the target description, augmented with~four task embeddings, is comprehensively interpreted into a set of diversified goal vectors, which are refined as the navigation progresses, and used as queries to retrieve supportive context from episodic history for decision making. This enables the reuse of knowledge across navigation tasks with varying input domains/modalities. We empirically demonstrate that, compared with learning each visual navigation task individually, our multitask agent achieves comparable or even better performance with reduced complexity.
\end{abstract}
\section{Introduction}
\begin{figure}
	\begin{center}
		\includegraphics[width=\linewidth]{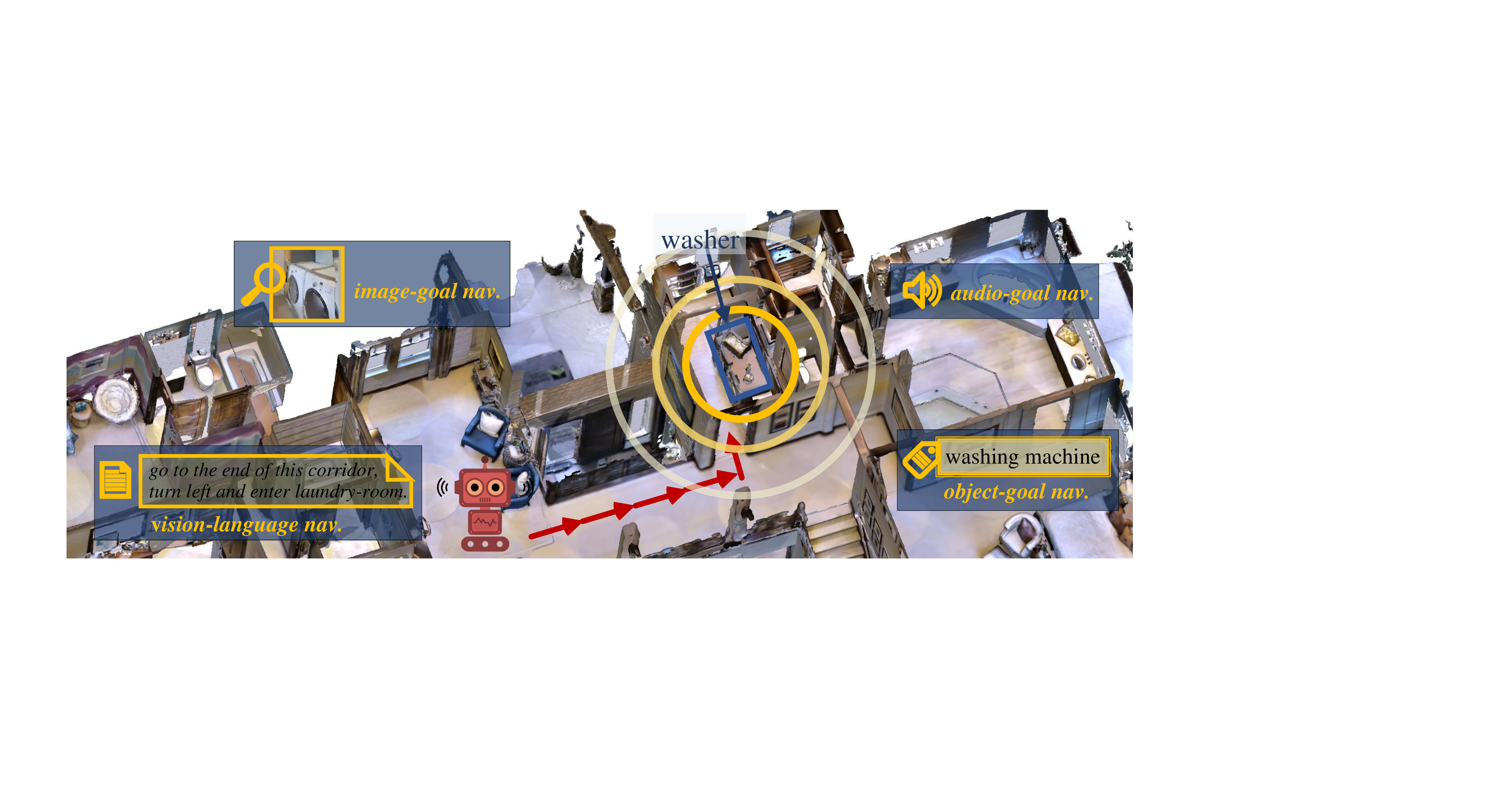}
	\end{center}
	\caption{{$_{\!}$Rather$_{\!}$ than$_{\!}$ existing$_{\!}$ efforts$_{\!}$ training$_{\!}$ specialized$_{\!}$ agents$_{\!}$ on$_{\!}$ individual$_{\!}$ navigation$_{\!}$ tasks, we build a single powerful agent that can undertake multiple tasks, \ie, \protect\includegraphics[scale=0.08,valign=c]{figs/image}$_{\!}$~\textit{image-goal nav.}, \protect\includegraphics[scale=0.1,valign=c]{figs/aud}$_{\!}$~\textit{audio-goal nav.}, \protect\includegraphics[scale=0.1,valign=c]{figs/tag}$_{\!}$~\textit{object-goal nav.}, and \protect\includegraphics[scale=0.1,valign=c]{figs/text}$_{\!}$~\textit{vision-language nav.}, in visually and acoustically realistic environments.}}
	\label{fig:topright}
	\vspace{-13pt}
\end{figure}

As a fundamental research topic, visual navigation has attained extensive attention across many~disciplines, including robotics$_{\!}$~\cite{kim1999symbolic}, computer vision$_{\!}$~\cite{chaplot2020object,chaplot2020neural}, and natural language processing$_{\!}$~\cite{striegnitz2011report}. Consider~a typical navigation scenario (Fig.$_{\!}$~\ref{fig:topright}), in which a human intends to direct a robot agent to navigate to a target -- a buzzing washer. The target can be specified by a photo of the washer (\ie, \protect\includegraphics[scale=0.1,valign=c]{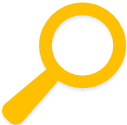}$_{\!}$~\textbf{\textit{image-goal nav.}}$_{\!}$~\cite{zhu2017target}), or the buzzing sound (\ie, \protect\includegraphics[scale=0.1,valign=c]{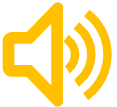}$_{\!}$~\textbf{\textit{audio-goal nav.}}$_{\!}$~\cite{chen2020soundspaces}), or the corresponding semantic~tag -- \textit{washing machine} (\ie, \protect\includegraphics[scale=0.1,valign=c]{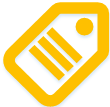}$_{\!}$~\textbf{\textit{object-goal nav.}}$_{\!}$~\cite{yang2018visual}), or linguistic instructions -- ``\textit{go~to~the end of this~corridor, turn left and enter the laundry-room}'' (\ie, \protect\includegraphics[scale=0.1,valign=c]{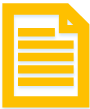}$_{\!}$~\textbf{\textit{vision-language nav.}}$_{\!}$~\cite{anderson2018vision}). Naturally, the agent~is expected to be smart enough to execute all these kinds of navigation tasks involving varying modalities/domains (\ie, image, audio, semantic tag, text) with different optimal policies. Contrary~to our expectation, almost all existing navigation agents are specifically designed/trained for one specific task -- a ``versatile'' agent capable of mastering multiple navigation tasks remains far beyond reach.

Besides its great value in practice, investigating embodied navigation in multitask scenarios can~help better understand human intelligence. First, we humans can learn multiple tasks in a parallel~ad~hoc manner, and benefit from commonalities across related tasks$_{\!}$~\cite{zhang2018overview}. Second, we accomplish tasks by processing and combining signals from different modalities. Evidences from cognitive psychology indicate that our senses are functioning together and multisensory integration is a central tenant~of human intelligence$_{\!}$~\cite{meredith1986visual,kording2007causal}. Though the idea of multitask learning$_{\!}$~\cite{caruana1997multitask} was widely explored~in~computer vision field$_{\!}$~\cite{crawshaw2020multi}, prior attempts are~often made in unsupervised and supervised learning settings; in the context of multitask reinforcement learning (MTRL)$_{\!}$~\cite{zhang2021survey}, not much is done for visually-rich, embodied navigation scenarios. One possible reason is the lack of a suitable dataset, compounded by considerable costs involved in data collection, as multiple navigation tasks should be supported.

In response, a large-scale 3D dataset, \textbf{\texttt{VXN}}, is established to investigate \textit{multitask multimodal embodied navigation} in \textit{audiovisual complex} indoor environments. \textbf{\texttt{VXN}} allows simulated robot agents to concurrently learn four tasks, \ie, \textit{image-goal~nav.}, \textit{audio-goal nav.}, \textit{object-goal nav.}, and~\textit{{vision-language nav.}}, in continuous, {acoustically-realistic}$_{\!}$~and$_{\!}$  {perceptually-rich}$_{\!}$ world\footnote{Strictly speaking, as we synthesize  visually and acoustically realistic environments, the classic \textit{image-goal nav.}, \textit{object-goal nav.}, and \textit{{vision-language nav.}} tasks in our \textbf{\texttt{VXN}} dataset are extended as \textit{image-goal visual-audio nav.}, \textit{visual-audio object-goal  nav.}, and \textit{visual-audio-language nav.}, respectively.}.$_{\!}$~Based$_{\!}$ on$_{\!}$ a$_{\!}$ high-throughput simulator$_{\!}$~\cite{2019Habitat}, \textbf{\texttt{VXN}} instantiates different navigation tasks in unified~environments following the same physical rules. It equips the agents with~multimodal sensors to gather information from 360$^{\circ\!}$ RGBD and audio observations. Taken all together, \textbf{\texttt{VXN}} provides a realistic testbed for multitask navigation.

With$_{\!}$ \texttt{VXN},$_{\!}$ we$_{\!}$ further$_{\!}$ develop$_{\!}$ \textsc{Vienna},$_{\!}$ a$_{\!}$ \underline{v}ersat\underline{i}le$_{\!}$~\underline{e}mbodied$_{\!}$ \underline{n}avigatio\underline{n}$_{\!}$ \underline{a}gent$_{\!}$ that$_{\!}$ jointly$_{\!}$ learns$_{\!}$ to solve the~four navigation tasks using \textit{one single model} without switching among different models.  Based on Transformer encoder-decoder architecture$_{\!}$~\cite{vaswani2017attention}, \textsc{Vienna} encodes the full episode history of multisensory inputs (\ie, RGB, depth, and audio) and navigation actions, and absorbs common knowledge across different navigation tasks with a shared decoder. Target signals (\ie, goal picture, target class, aural cues, linguistic instruction) are parsed into \textit{queries}, and the supportive context retrieved from the encoded history is fed to corresponding  \textit{policy} for task-specific decision making. 
With such a \textit{fully-attentive} model design, \textsc{Vienna} is able to comprehend multimodal observations, conduct long-term reasoning, and, more essentially, exploit cross-task knowledge.

By contrasting our \textsc{Vienna} to several single-task counterparts on \textbf{\texttt{VXN}}, we empirically demonstrate \textbf{i)}~\textit{Better performance}.$_{\!}$ Through$_{\!}$ exploiting$_{\!}$ cross-task$_{\!}$ relatedness, \textsc{Vienna} outperforms independent task training.  \textbf{ii)}~\textit{Reduced model-size}. Training four tasks together using a single \textsc{Vienna} achieves about four times model size compression, compared with training them individually. \textbf{iii)}~\textit{Improved generalization}. \textsc{Vienna} performs robust on unseen environments, through learning task-shared, general representations. \textbf{iv)}~\textit{More is better}. The above conclusions are typically true when we train \textsc{Vienna} on more navigation tasks. \textbf{v)}~\textit{Multisensory$_{\!}$ integration$_{\!}$ does$_{\!}$ matter}.$_{\!}$ Both$_{\!}$ visual (RGB and depth) and aural information are crucial building blocks for general-purpose navigation robot creation.

\section{Related Work}
\noindent\textbf{Embodied Navigation.} As a fundamental element in building intelligent robots, navigation has long been the focus of the scientific community$_{\!}$~\cite{giralt1979multi}. The availability of building-scale 3D datasets$_{\!}$~\cite{armeni20163d,chang2017matterport3d,wu2018building,song2017semantic} and high-performance simulation platforms$_{\!}$~\cite{kolve2017ai2,2019Habitat,xia2018gibson,deitke2020robothor} led to a plethora of reproducible research of navigation in large-scale, visually-rich environments. Depending on how to specify the target~goal, diverse navigation tasks are proposed to let an agent \textbf{i)} navigate to target coordinates (\textit{point-goal nav.}$_{\!}$~\cite{2019Habitat}), \textbf{ii)}  find an instance of a given object category (\textit{object-goal nav.}$_{\!}$~\cite{yang2018visual}), \textbf{iii)}  search for~target$_{\!}$ photos$_{\!}$ (\textit{image-goal$_{\!}$ nav.}$_{\!}$~\cite{zhu2017target}),$_{\!}$ \textbf{iv)}$_{\!}$  locate$_{\!}$ sound$_{\!}$ sources$_{\!}$ (\textit{audio-goal$_{\!}$ nav.}$_{\!}$~\cite{chen2020soundspaces}),$_{\!}$ or$_{\!}$ \textbf{v)}$_{\!}$  follow$_{\!}$ navigation$_{\!}$~instru- ctions (\textit{vision-language$_{\!}$ nav.}$_{\!}$~\cite{anderson2018vision,krantz2020navgraph}).$_{\!}$ Aside from these battlefields, there are some$_{\!}$ more$_{\!}$~complicated embodied$_{\!}$ tasks,$_{\!}$ such$_{\!}$ as$_{\!}$ \textit{embodied question answering}$_{\!}$~\cite{das2018embodied}, \textit{vision-dialog nav.}$_{\!}$~\cite{2019Vision,2018Vision,2019Help},  and \textit{multiagent$_{\!}$ nav.}$_{\!}$~\cite{wang2021collaborative}.
The$_{\!}$ community$_{\!}$ also$_{\!}$ made$_{\!}$ great$_{\!}$ strides$_{\!}$ in$_{\!}$ improving$_{\!}$ reinforcement$_{\!}$ learning$_{\!}$ (RL)$_{\!}$ algorithms capable of fulfilling specific navigation tasks, by using, for example, recurrent~neural networks$_{\!}$~\cite{mirowski2017learning,2019Habitat,anderson2018vision}, map building \cite{gupta2017cognitive,parisotto2017neural,zhang2017neural,wu2019bayesian,chaplot2018active,savinov2018semi,chaplot2020neural,chaplot2020learning,wang2021structured,Chen_2021_CVPR,zhao2022target}, path planning$_{\!}$~\cite{lee2018gated,deng2020evolving,wang2020active}, cross-modal attention$_{\!}$~\cite{hu2019you,qi2020object,wang2020environment,Hong_2021_CVPR,zhao2022target}, synthesized or unlabeled data$_{\!}$~\cite{fried2018speaker,tan2019learning,fu2019counterfactual,majumdar2020improving,hao2020prevalent,wang2022counterfactual}, and external knowledge$_{\!}$~\cite{yang2018visual,gao2021room}.
However, though the learning algorithm is general$_{\!}$ --$_{\!}$ RL,$_{\!}$ each$_{\!}$ solution$_{\!}$ is$_{\!}$ not;$_{\!}$ each$_{\!}$ navigation$_{\!}$ agent$_{\!}$ can$_{\!}$ only$_{\!}$ handle$_{\!}$ the$_{\!}$ one$_{\!}$ task$_{\!}$ it$_{\!}$ was$_{\!}$ trained$_{\!}$ on.

$_{\!}$With$_{\!}$ various$_{\!}$ navigation$_{\!}$ tasks$_{\!}$ and$_{\!}$ task-specific$_{\!}$ navigation$_{\!}$ solutions, a critical question arises \textit{whether we can build a single general agent that works~well for multiple navigation tasks}. In~response, we make$_{\!}$ two$_{\!}$ unique$_{\!}$ contributions.$_{\!}$ First,$_{\!}$ we$_{\!}$ build$_{\!}$ a$_{\!}$ large-scale$_{\!}$ 3D$_{\!}$ dataset$_{\!}$ that$_{\!}$ supports$_{\!}$ four$_{\!}$ representative navigation tasks in continuous and realistic environments. In contrast, prior navigation datasets are built upon different platforms and with certain assumptions/configurations (\eg, sparse navigation graphs$_{\!}$~\cite{anderson2018vision}, discrete world representation$_{\!}$~\cite{chen2020soundspaces}),  making them hard to explore different navigation~tasks~in unified and standardized environments. Second, we create a generalist agent which is capable of undertaking a set of navigation tasks of different modalities/domains, and is equipped with multimodal sensors (\ie, RGB, depth, audio) to better address real-world scenarios. However, existing navigation agents are trained one task at the time, each new task requiring to train a new agent instance.

\noindent\textbf{Multitask$_{\!}$ Learning$_{\!}$ (MTL).}$_{\!}$ MTL$_{\!}$~\cite{caruana1997multitask}, inspired$_{\!}$ by$_{\!}$~the human ability to$_{\!}$ transfer$_{\!}$ knowledge across different tasks$_{\!}$~\cite{ruder2017overview}, has led to wide success in computer vision$_{\!}$~\cite{wang2018attentive,wang2019salient,lu202012}  and natural language pro- cessing$_{\!}$~\cite{collobert2008unified}. Related efforts were made along three directions$_{\!}$~\cite{crawshaw2020multi}: \textbf{i)} \textit{architecture design} (\ie, how to partition the model into~task-specific$_{\!}$ and$_{\!}$ shared$_{\!}$ components)$_{\!}$~\cite{zhang2014facial,misra2016cross,xu2018pad,strezoski2019many}, \textbf{ii)} \textit{optimization} (\ie, how$_{\!}$ to$_{\!}$ balance$_{\!}$ learning$_{\!}$ between$_{\!}$ different$_{\!}$ tasks) \cite{kendall2018multi,chen2018gradnorm,guo2018dynamic,duong2015low,sanh2019hierarchical,sener2018multi}, and \textbf{iii)} \textit{task relationship learning} (\ie, how to learn and utilize task relationships to improve learning) \cite{bingel2017identifying,zamir2018taskonomy,dwivedi2019representation}. In the field of MTRL$_{\!}$~\cite{yang2017multi,teh2017distral,espeholt2018impala,pinto2017learning,zeng2018learning,hessel2019multi,d2020sharing}, recent solutions explored knowledge transfer$_{\!}$~\cite{xu2020knowledge}, modular networks$_{\!}$~\cite{heess2016learning,devin2017learning}, and policy distillation$_{\!}$~\cite{rusu2015policy,parisotto2016actor}. A few robotics benchmarks$_{\!}$~\cite{beattie2016deepmind,bellemare2013arcade,yu2020meta} are also proposed for MTRL. However, most of these efforts were based upon low-dimension observations, \eg, grid-world like or game environments. To the best of our knowledge, there are two prior work$_{\!}$~\cite{wang2020environment,chaplot2019embodied} that addressed  multitask navigation, but they only consider two closely-related, language-guided navigation tasks with the same input modalities.

$_{\!}$Drawing$_{\!}$ inspiration$_{\!}$ from$_{\!}$ these$_{\!}$ efforts,$_{\!}$ we$_{\!}$ seek$_{\!}$ for a ``universal'' agent that can complete multiple navigation tasks with a single agent instance, and distinguish ourselves by \textbf{i)} joint learning of four navigation tasks with diverse input modalities, \textbf{ii)} visually complex and acoustically realistic operation space, \textbf{iii)} multisensory integration, and \textbf{iv)} fully-attentive architecture based parse-and-query regime.

\noindent\textbf{Auxiliary$_{\!}$ Learning$_{\!}$ in$_{\!}$ Embodied$_{\!}$ Navigation.}$_{\!}$~There$_{\!}$ are$_{\!}$ a$_{\!}$ group$_{\!}$ of$_{\!}$ algorithms$_{\!}$ that$_{\!}$ exploit$_{\!}$ comple- mentary objectives from auxiliary tasks to facilitate navigation policy learning. Specifically, \textit{supervised auxiliary tasks} expose privileged information to the agent~(\eg,~depth$_{\!}$~\cite{mirowski2016learning}, surface normals$_{\!}$~\cite{gordon2019splitnet}, semantics$_{\!}$~\cite{das2018embodied}, \etc). \textit{Self-supervised auxiliary tasks} derive free supervisory signals from the agent's own experience (\eg, next-step visual feature prediction$_{\!}$~\cite{pathak2017curiosity}, predictive modeling$_{\!}$~\cite{gregor2019shaping}, loop closure prediction$_{\!}$~\cite{mirowski2016learning}, temporal distance estimation$_{\!}$~\cite{ye2020auxiliary}, navigation progress estimation$_{\!}$~\cite{ma2019self,zhu2019vision}, \etc).

Although auxiliary learning based navigation models are also trained on a set of tasks, their ideas are far away from ours.
These models still focus on only a single ``main'' navigation task with extra aid of auxiliary intermediate objectives, while we aim to capture and utilize common knowledge of a collection of different navigation tasks to enhance the performance on all the tasks. Moreover, their auxiliary tasks, in principle, can be utilized by our agent, but they cannot handle our task setting.

\noindent\textbf{Transformer in Embodied Navigation and MTL.} Inspired by the great success of Transformer$_{\!}$~\cite{vaswani2017attention} in$_{\!}$ sequence$_{\!}$ transduction$_{\!}$ tasks,$_{\!}$ 
a$_{\!}$ few$_{\!}$ recent$_{\!}$ methods$_{\!}$ applied$_{\!}$ Transformer for certain navigation tasks$_{\!}$~\cite{fang2019scene,landi2020perceive,chen2021semantic,du2021vtnet,Chen_2021_CVPR,pashevich2021episodic}. Rather than sharing similar advantages in long-term memory and cross-modal information fusion, our method further formulates different navigation tasks as a unified process of active goal parsing and supportive information query. Through cleverly encoding all task-specific embeddings into goal parsing, our agent is able to explicitly leverage cross-task knowledge to boost different navigation tasks. There are also a few notable studies that exploit Transformer-like network architectures for MTL$_{\!}$~\cite{kaiser2017one,lu2019vilbert,pramanik2019omninet,hu2021unit}, while none of them addresses embodied visual tasks.

\noindent\textbf{Pretraining in Embodied Navigation.} A series of methods decompose embodied tasks into visual (and linguistic) representation learning and policy training$_{\!}$~\cite{gordon2019splitnet,majumdar2020improving,hao2020towards,Li_2020_CVPR,guhur2021airbert}. They pretrain a general model on easily-acquired viual or multimodal data (\eg, image captions) and fine-tune the policy for ``downstream'' navigation tasks. Though showing improved generalization and transfer abilities, the result is still a collection of independent task-specific models rather than a single agent instance.

\section{\textbf{\texttt{VXN}} Dataset for Multitask Multimodal Embodied Navigation}

\subsection{Task Collection and Dataset Acquisition}
\textbf{\texttt{VXN}} includes four famous navigation tasks, \ie,$_{\!}$ \textit{image-goal$_{\!}$ nav.}$_{\!}$~\cite{zhu2017target}, \textit{audio-goal$_{\!}$ nav.}$_{\!}$~\cite{chen2020soundspaces}, \textit{object-goal$_{\!}$ nav.}$_{\!}$~\cite{yang2018visual}, and$_{\!}$ \textit{vision-language$_{\!}$ nav.}$_{\!}$~\cite{anderson2018vision}.$_{\!}$ These$_{\!}$ tasks$_{\!}$ are$_{\!}$ with different$_{\!}$ input$_{\!}$ modalities/domains$_{\!}$ (\ie, visual,$_{\!}$ audio,$_{\!}$ semantic~tag, and language); their original datasets adopt  different world representations$_{\!}$  (\ie,$_{\!}$ graph$_{\!}$  based$_{\!}$~\cite{anderson2018vision} \textit{vs}~discrete$_{\!}$~\cite{chen2020soundspaces} \textit{vs}~continuous$_{\!}$~\cite{2019Habitat}), environment configurations (\ie, visually poor$_{\!}$~\cite{kolve2017ai2}~\textit{vs} perception rich~\cite{anderson2018vision} \textit{vs}~audiovisual realistic$_{\!}$~\cite{chen2020soundspaces}),  and success criteria (\ie, 3 m~\cite{anderson2018vision} \textit{vs} 1~m$_{\!}$~\cite{zhu2017target} \textit{vs} 1 m$_{\!}$~\cite{yang2018visual} \textit{vs} 1 m$_{\!}$~\cite{chen2021semantic}). Hence, to study these four tasks in a single learning system, it is desired to build a standardized dataset that initiates them with similar problem settings, \eg, dynamic transition, world representations, and audiovisual properties, instead of simply combining several single-task navigation datasets together. On the other hand, it is wise to maximize the reuse of existing  datasets, ensuring continuity and compatibility \wrt~former research, and reducing data annotation cost.

As$_{\!}$ many$_{\!}$ previous$_{\!}$ navigation$_{\!}$ datasets$_{\!}$ \cite{anderson2018vision,chen2020soundspaces,batra2020objectnav}$_{\!}$ are$_{\!}$ built$_{\!}$ upon$_{\!}$ Matterport3D (MP3D)$_{\!}$~\cite{chang2017matterport3d} environments and Habitat$_{\!}$~\cite{2019Habitat}  simulator,$_{\!}$ we$_{\!}$ derive$_{\!}$ a$_{\!}$ unified,$_{\!}$ multitask$_{\!}$ navigation$_{\!}$ dataset -- \texttt{VXN} --  by$_{\!}$ converting$_{\!}$ previous$_{\!}$ \textit{task-specific}$_{\!}$ datasets$_{\!}$ to$_{\!}$ \textit{standardized, continuous,$_{\!}$ audiovisual-rich$_{\!}$ environments}:
\begin{itemize}[leftmargin=*]
	\setlength{\itemsep}{0pt}
	\setlength{\parsep}{-0pt}
	\setlength{\parskip}{-0pt}
	\setlength{\leftmargin}{-8pt}
    \item Our \textbf{\textit{audio-goal nav.}}~is built upon SoundSpaces$_{\!}$~\cite{chen2020soundspaces}, which offers audio renderings for MP3D~and allows to navigate sounding targets, or conduct point navigation with extra aid of audio cues.~Due~to heavy$_{\!}$ acoustic$_{\!}$ simulation$_{\!}$ cost,$_{\!}$ \cite{chen2020soundspaces}$_{\!}$ uses$_{\!}$ a$_{\!}$ grid-based$_{\!}$ world$_{\!}$ model:$_{\!}$ it$_{\!}$ samples$_{\!}$ room$_{\!}$ impulse$_{\!}$~res- ponse$_{\!}$ over$_{\!}$ a$_{\!}$ discrete,$_{\!}$ horizontal$_{\!}$ plane$_{\!}$ (1.5~$_{\!}$m$_{\!}$ above$_{\!}$ the$_{\!}$ floor$_{\!}$ with$_{\!}$ 0.5~$_{\!}$m$\times$0.5~$_{\!}$m$_{\!}$ grid$_{\!}$ size).$_{\!}$ We$_{\!}$ devise an~audio$_{\!}$ simulator$_{\!}$ to$_{\!}$ efficiently$_{\!}$ transfer$_{\!}$ grid-level$_{\!}$ audio$_{\!}$ renderings$_{\!}$ into$_{\!}$ continuous$_{\!}$ setting$_{\!}$ (\textit{cf}.$_{\!}$~\S\ref{sec:dd}).
    \item Our \textbf{\textit{vision-language$_{\!}$ nav.}}~is built upon R2R$_{\!}$~\cite{anderson2018vision}, which labels MP3D with linguistic navigation~ins- tructions.$_{\!}$ R2R$_{\!}$ is$_{\!}$ yet$_{\!}$ bounded$_{\!}$ to$_{\!}$ graph-based$_{\!}$ world$_{\!}$ representation$_{\!}$ -- each scene$_{\!}$ can$_{\!}$ be$_{\!}$ only$_{\!}$~observed$_{\!}$ from$_{\!}$ a$_{\!}$ few$_{\!}$ fixed$_{\!}$ points$_{\!}$ ($\sim$$_{\!\!}$~117) and environment topologies are pre-given. We use \cite{krantz2020navgraph} to convert R2R to the continuous setting, and then adopt~\cite{chen2020soundspaces} and our audio simulator for audio rendering.
    \item Our \textbf{\textit{image-goal$_{\!}$ nav.}}~is built upon Habitat$_{\!}$~\cite{2019Habitat} ImageNav repository$_{\!}$~\cite{habitat}, which is for photo target guided navigation in MP3D environments. Again, continuous audio rendering is made.
    \item Our \textbf{\textit{object-goal$_{\!}$ nav.}}~is built upon Habitat2020 ObjectNav challenge$_{\!}$~\cite{batra2020objectnav}, which requires~an~agent to navigate a MP3D environment to find an instance of an object class. A total of 21 visually well defined object categories (\eg, \textit{chair}) are considered and audio rendering is also made; but the GPS + Compass sensor, used in$_{\!}$~\cite{batra2020objectnav}, is not adopted in \texttt{VXN}, for formalizing different task settings.
\end{itemize}

\subsection{Task Setting and Dataset Design}\label{sec:dd}

\begin{figure*}
    \begin{minipage}{\textwidth}
        \begin{minipage}[t]{0.54\textwidth}
            \makeatletter\def\@captype{table}\captionsetup{width=.8\linewidth}
            \caption{Data splits and the number of navigation episodes in our \textbf{\texttt{VXN}} dataset (\S\ref{sec:dd}).}
            \label{tab:dataset}
            \centering\small
            \resizebox{1.\linewidth}{!}{
                \setlength\tabcolsep{2pt}
                \renewcommand\arraystretch{1.2}
                \begin{tabular}{|c|rlrlrl|}
			\hline
            Navigation Task &\multicolumn{2}{c}{\tabincell{c}{\texttt{train}\\(58 environments)}} & \multicolumn{2}{c}{\tabincell{c}{\texttt{val seen}\\(58 environments)}}  & \multicolumn{2}{c|}{\tabincell{c}{\texttt{val unseen}\\(11 environments)}} \\\hline
            \textit{Audio-goal} &2.0M &episodes  &500 &episodes  &500 &episodes \\
            \textit{Vision-language} &10,819 &episodes  & 778 &episodes  & 1,839 &episodes\\
            \textit{Object-goal}  &2.6M &episodes &{500} &episodes  &2,195 &episodes \\
            \textit{Image-goal} &5.0M &episodes &495 &episodes   &495 &episodes \\
            \hline
            {Total} &9.6M &episodes &{2,273} &episodes   &5,029 &episodes \\
            \hline
		\end{tabular}
            }
        \end{minipage}
        \hspace{-0.1ex}
        \begin{minipage}[t]{0.45\textwidth}
            \makeatletter\def\@captype{table}\captionsetup{width=.94\linewidth}
            \caption{Comparison (\S\ref{sec:dd}) of  navigation  datasets \footnotesize{(\text{MT}: multitask; \texttt{CS}: continuous space; \texttt{VR/AR}: visual/audio realistic; \texttt{PA}: panorama).}}
	\label{tab:comparison}
            \centering\small
            \resizebox{1.\linewidth}{!}{
                \setlength\tabcolsep{7pt}
                \renewcommand\arraystretch{1.}
                \begin{tabular}{|c|c|ccccc|}
    			\hline
                Navigation Dataset &Year & \text{MT}  & \texttt{CS} & \texttt{VR} & \texttt{AR} & \texttt{PA} \\\hline
                EQA~\cite{das2018embodied} &2018 & &  & &  &   \\
                Habitat-PointGoal~\cite{2019Habitat} &2019 & &\checkmark  &\checkmark &  &   \\
                R2R~\cite{anderson2018vision}  &2018 & &  &\checkmark &  &   \\
                VLN-CE~\cite{krantz2020navgraph}  &2020 & &\checkmark  &\checkmark &  &   \\
                Gibson-ImageGoal~\cite{chaplot2020neural} &2020 & &\checkmark  &\checkmark &  &\checkmark   \\
                SoundSpaces~\cite{chen2020soundspaces}  &2020 & &  &\checkmark &\checkmark  &  \\
                \hline
                \textbf{\texttt{VXN}} &2022 & \checkmark  & \checkmark  & \checkmark & \checkmark & \checkmark \\\hline
        		\end{tabular}
            }
        \end{minipage}
    \end{minipage}
\end{figure*}

\noindent\textbf{Panoramic Visual Simulator.$_{\!}$} With Habitat API, 360$^{\circ\!}$~egocentric RGBD view is rendered at 300~fps.

\noindent\textbf{Audio Simulator.} With$_{\!}$~\cite{chen2020soundspaces}, ambisonics are generated at locations sampled in MP3D scenes~and converted$_{\!}$ to$_{\!}$ binaural$_{\!}$ audio$_{\!}$~\cite{2018Binaural},$_{\!}$ \ie,$_{\!}$ an$_{\!}$ agent$_{\!}$ emulates$_{\!}$ two$_{\!}$ human-like$_{\!}$ ears.$_{\!}$ To$_{\!}$ synthesize$_{\!}$ continuous$_{\!}$ auditory$_{\!}$ scenes,$_{\!}$ we$_{\!}$ use$_{\!}$~\cite{kearney2009approximation} for real-time binaural room impulse responses (BRIRs) interpolation. We adopt Dynamic Time Wrapping$_{\!}$~\cite{sakoe1978dynamic} to temporally align left and right ear BRIRs and~then map the warped \textit{interpolated} vectors back into the ``unwarped'' time domain, to get BRIRs at arbitrary locations  and directions. As in$_{\!}$~\cite{chen2021semantic}, the \href{https://github.com/facebookresearch/sound-spaces/}{sounds} of the 21 object categories$_{\!}$~\cite{batra2020objectnav} in \textit{object-goal~nav.}~are used for audio rendering. Moreover, the sounds are associated with the objects of same semantic categories to ensure generating semantically meaningful and contextual audio$_{\!}$~\cite{chen2021semantic}.  For \textit{image-goal$_{\!}$ nav.}, \textit{object-goal$_{\!}$ nav.}, and$_{\!}$ \textit{vision-language$_{\!}$ nav.}, the audio is used as background sound, which can reveal the geometry of environment$_{\!}$~\cite{chen2020soundspaces}, complement the visual cues, and make the tasks closer to the real-world. For \textit{audio-goal$_{\!}$ nav.}, the navigation target is directly specified by the audio.

\noindent\textbf{Episodes and Dataset Splits.} In \textbf{\texttt{VXN}}, each episode is defined as a tuple:$_{\!}$ $\langle$scene, audio waveform, agent start location, agent start$_{\!}$ rotation,$_{\!}$ goal$_{\!}$ location,$_{\!}$ target$_{\!}$ description$\rangle$.$_{\!}$ We$_{\!}$ use$_{\!}$ the standard 58/11/18 \texttt{train}/\texttt{val}/\texttt{test} split$_{\!}$~\cite{anderson2018evaluation} of MP3D environments. Since previous navigation datasets~\cite{anderson2018vision,batra2020objectnav,2019Habitat} keep \texttt{test} annotations private, we only use \texttt{train} and \texttt{val} environments to create \textbf{\texttt{VXN}} (\textit{cf}.~Table$_{\!}$~\ref{tab:dataset}). 

\noindent\textbf{Action Space.} We adopt a panoramic action space, which is widely used in recent embodied robotic tasks$_{\!}$~\cite{chaplot2020neural, deng2020evolving}. Specifically, the panoramic view is horizontally discreted into a total of 12 sub-views. Agents can move towards a sub-view 0.25 m or \textsf{stop}.

\noindent\textbf{Success Criterion.}  An episode is considered as successful if
the agent \textbf{i)} executes \textsf{stop} action,  \textbf{ii)} within 1 m of
the goal location, and \textbf{iii)} within a time horizon of 500 actions (as in$_{\!}$~\cite{chen2019touchdown,gordon2019splitnet,2019Habitat}).

\noindent\textbf{Dataset Features.} As shown in Table$_{\!}$~\ref{tab:comparison}, \textbf{\texttt{VXN}} poses greater challenges: the agent needs to master four navigation tasks with various input modalities in continuous, audiovisual complex environments,~mine cross-task knowledge, and reason intelligently about all the senses available to it (RGB, depth,~audio).

\section{Our Approach} \label{sec:approach}

\noindent\textbf{Problem$_{\!}$ Statement.$_{\!}$} In$_{\!}$ \textit{single-task$_{\!}$ navigation},$_{\!}$ an$_{\!}$ {agent}$_{\!}$ learns$_{\!}$ to$_{\!}$ reach$_{\!}$ a$_{\!}$ goal$_{\!}$ position.$_{\!}$ This$_{\!}$ is$_{\!}$~typically formulated$_{\!}$ in$_{\!}$ a$_{\!}$ RL$_{\!}$ framework$_{\!}$ that$_{\!}$ solves$_{\!}$ a$_{\!}$ partially$_{\!}$ observable$_{\!}$ Markov$_{\!}$ decision$_{\!}$ process$_{\!}$~\cite{bellman1957markovian}:$_{\!}$ a$_{\!}$ tuple $(\mathcal{S},\mathcal{A},\mathcal{G},{O},P, R, \gamma)$, where $\mathcal{S}$, $\mathcal{A}$, $\mathcal{G}$ are sets of \textit{states}, \textit{actions} and \textit{targets}, $o_{t\!}=_{\!}O(s_t)$~denotes the local observation at global state $s_{t\!}\in_{\!}\mathcal{S}$ at epoch (decision step) $t$, $P(s_{t+1\!}|s_t,a_t)$ is the transition~probability from$_{\!}$ $s_{t\!}$ to$_{\!}$ $s_{t+1\!}$ given$_{\!}$ action$_{\!}$ $a_{t\!}\!\in_{\!}\!\mathcal{A}$,$_{\!}$ $R(s,a)_{\!}\!\in_{\!}\!\mathbb{R}$$_{\!}$ gives$_{\!}$ the$_{\!}$ \textit{reward}, and $\gamma\!\in\!(0,1)$ discounts future rewards.$_{\!}$ The$_{\!}$ agent$_{\!}$ uses$_{\!}$ a$_{\!}$ \textit{policy}$_{\!}$ $\pi(a|o, g)$ to produce its action $a$, conditioned on its local observation $o$ and target goal $g\!\in_{\!}\mathcal{G}$, and optimizes~its accumulated discounted reward $J_{\!\!}=_{\!\!}\sum_{t'=t\!}^T\gamma^{t'\!-t\!}R(s_{t'},a_{t'})$.

In$_{\!}$ our$_{\!}$ \textit{multitask$_{\!}$ navigation},$_{\!}$ a$_{\!}$ single$_{\!}$ agent$_{\!}$ needs$_{\!}$~to$_{\!}$ master$_{\!}$ $K_{\!}\!=_{\!}\!4$ tasks,$_{\!}$ \ie,$_{\!}$ \{\textit{audio-goal},$_{\!}$ \textit{object-goal}, \textit{image-goal},$_{\!}$ \textit{vision-language}\}$_{\!}$ in$_{\!}$ \textbf{\texttt{VXN}}$_{\!}$ environments.$_{\!}$ We$_{\!}$ formalize$_{\!}$ this$_{\!}$ as$_{\!}$ a$_{\!}$ MTRL$_{\!}$ problem:$_{\!\!}$~$\{(\mathcal{S},\mathcal{A},~\mathcal{G}_k,$ ${O},P,R_k,\gamma_k)\}_{k=1}^K$, where$_{\!}$ the$_{\!}$ agent$_{\!}$ concurrently$_{\!}$ learns$_{\!}$ $K$$_{\!}$ task-specific$_{\!}$ policies$_{\!}$ $\pi_{1:K\!}$ that maximize~the rewards$_{\!}$ $J_{1:K}$.$_{\!}$ The$_{\!}$ single$_{\!}$ multitask$_{\!}$ agent$_{\!}$ is$_{\!}$ expected$_{\!}$ to$_{\!}$ exploit$_{\!}$ cross-task$_{\!}$ knowledge$_{\!}$ to$_{\!}$ achieve$_{\!}$ close$_{\!}$~or better$_{\!}$ navigation$_{\!}$ performance$_{\!}$ on$_{\!}$ the$_{\!}$ $K_{\!}$ tasks,$_{\!}$ compared$_{\!}$ with$_{\!}$ training$_{\!}$ $K_{\!}$ single-task$_{\!}$ agents$_{\!}$ individually.

\noindent\textbf{Transformer$_{\!}$ Preliminary.$_{\!}$} The$_{\!}$ core$_{\!}$ of$_{\!}$ Transformer$_{\!}$~\cite{vaswani2017attention} is$_{\!}$ an$_{\!}$ attention$_{\!}$ function$_{\!}$ (denoted$_{\!}$ as$_{\!}$ ${f}_{\textsc{Att}}$), which takes a query sequence $\bm{x}_{\!}\!\in\!\mathbb{R}^{n_{\!}\times_{\!}d\!}$ and a context sequence $\bm{y}\!\in\!\mathbb{R}^{m_{\!}\times_{\!}d}$  as inputs, and outputs:
\begin{equation}\label{equ:attn}
	\!\!\tilde{\bm{y}}\! = \!{f}_{\textsc{Att}}(\bm{x}, \bm{y})\!= \! \text{softmax}\big((\bm{x}\bm{W}^{q})(\bm{y}\bm{W}^{k})^{\top\!}/\sqrt{d}\!~\big)\big(\bm{y}\bm{W}^{v}\big).
\end{equation}
where $\tilde{\bm{y}}\!\in\!\mathbb{R}^{n\times d\!}$ is with the same length $n$ and embedding dimension $d$ as $\bm{x}$, and$_{\!}$ $\bm{W}^{q,k,v\!}\!\in_{\!}\!\mathbb{R}^{d_{\!}\times_{\!}d\!}$ are learnable \textit{query}, \textit{key}, and \textit{value} projection matrices, respectively. Note that Eq.~\ref{equ:attn} is applicable to both \textit{self-attention} in Transformer encoder (\ie, $\bm{x}\!\equiv\!\bm{y}$), and \textit{cross-attention} in Transformer decoder (\ie, $\bm{x}\!\neq\!\bm{y}$). Further, each Transformer layer block can be given as:
\begin{equation}\label{equ:tb}
		\bm{x}'\!=\!\bm{x}\!+\!{f}_{\textsc{Mha}}(\bm{x}, \bm{y})\in\!\mathbb{R}^{n\times d},~~~~~~\bm{z}\!=\!\bm{x}'\!+\!{f}_{\textsc{Mlp}}(\bm{x}')\in\!\mathbb{R}^{n\times d},
\end{equation}
where ${f}_{\textsc{Mha}}$ refers to a \textit{multi-head attention} layer, derived by computing  several ${f}_{\textsc{Att}}$ in parallel, and ${f}_{\textsc{Mlp}}$ is multi-layer perceptron. The layer normalization is omitted for brevity.

\noindent\textbf{Core Idea.} Built upon a Transformer encoder-decoder architecture, our \textsc{Vienna} unifies the four~\texttt{VXN} tasks as an attention-based, \textit{parse-and-query} framework: the target description $g\!\in\!\mathcal{G}_k$ is \textit{online}~parsed into a set~of~embeddings, which are used to ``query'' the encoded episode history; the retrieved~suppor- tive cues are fed into the corresponding policy $\pi_k$ for decision making. To better handle multiple tasks, \textsc{Vienna} \textbf{i)} learns task-wise context and involves all the task-specific embeddings into target parsing, \textbf{ii)} shares representations among tasks, \textbf{iii)} lets task-specific policies $\pi_{1:K}$ reuse knowledge, and \textbf{iv)} trains the polices via a multitask version of Distributed Proximal Policy Optimization (DPPO)$_{\!}$~\cite{heess2017emergence}.

\begin{figure*}
	\begin{center}
		\includegraphics[width=\linewidth]{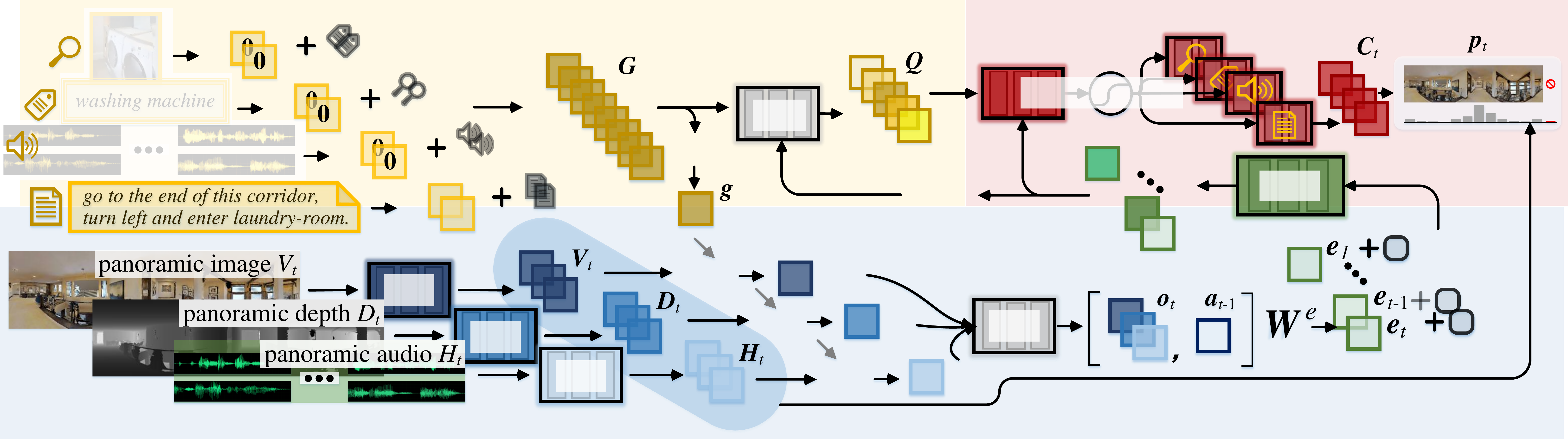}
		\put(-375,97){\tiny $g_\text{I}\!=\!\varnothing$}
        \put(-373,84){\tiny $g_\text{T}\!=\!\varnothing$}
        \put(-371,72){\tiny $g_\text{A}\!=\!\varnothing$}
        \put(-398,57){\tiny $g_\text{L}$}
        \put(-330,100){\tiny $\bm{g}_\text{I}$}
        \put(-314,85){\tiny $\bm{g}_\text{T}$}
        \put(-297,75){\tiny $\bm{g}_\text{A}$}
        \put(-283,60){\tiny $\bm{g}_\text{L}$}
        \put(-305,100){\tiny $\bm{\tau}_\text{I}$}
        \put(-288,85){\tiny $\bm{\tau}_\text{T}$}
        \put(-270,75){\tiny $\bm{\tau}_\text{A}$}
        \put(-255,60){\tiny $\bm{\tau}_\text{L}$}
        \put(-110,70){\tiny $\tilde{\bm{e}}_1$}
        \put(-102,60){\tiny $\tilde{\bm{e}}_{t-\!1}$}
        \put(-96,50){\tiny $\tilde{\bm{e}}_{t}$}
        \put(-37,46){\tiny $\bm{\mu}_1$}
        \put(-26,34){\tiny $\bm{\mu}_{t-\!1}$}
        \put(-22,27){\tiny $\bm{\mu}_{t}$}
        \put(-190,38){\tiny $\tilde{\bm{v}}_t$}
        \put(-174,28){\tiny $\tilde{\bm{d}}_t$}
        \put(-156,13){\tiny $\tilde{\bm{h}}_t$}
        \put(-260,104){\small Target Parser~(\S\ref{sec:TP})}
        \put(-140,104){\small Multitask Planner~(\S\ref{sec:MP})}
        \put(-95,3){\small Episodic Encoder~(\S\ref{sec:ME})}
        \put(-207,80){\tiny $f_{\textsc{Mha}}$}
        \put(-122,86){\tiny $f_{\textsc{Mtp}}$}
        \put(-77,62){\tiny $f_{\textsc{Ehe}}$}
        \put(-147,27){\tiny $f_{\textsc{Msi}}$}
        \put(-300,36){\tiny $f_{\textsc{Img}}$}
        \put(-278,24){\tiny $f_{\textsc{Dep}}$}
        \put(-257,14){\tiny $f_{\textsc{Aud}}$}
        \put(-190,15){\tiny $f_{\textsc{Att}}$}
        \put(-206,28){\tiny $f_{\textsc{Att}}$}
        \put(-223,41){\tiny $f_{\textsc{Att}}$}
        \put(-225,72){\tiny $f_{\textsc{Avg}}$}
        \put(-163,59){\tiny $f_{\textsc{Avg}}$}
	\end{center}
	\caption{{Detailed network architecture of \textsc{Vienna}, at epoch $t$ in a \textit{vision-language nav.} episode.}}
	\label{fig:pipe}
\end{figure*}

\textsc{Vienna}$_{\!}$ has$_{\!}$ three$_{\!}$ modules$_{\!}$ (\textit{cf}.$_{\!}$~Fig.$_{\!}$~\ref{fig:pipe}):$_{\!}$ \textbf{i)}$_{\!}$ an$_{\!}$ \textit{episodic}$_{\!}$ \textit{encoder}$_{\!}$ (\S\ref{sec:ME})$_{\!}$ that$_{\!}$ fuses$_{\!}$ multisensory$_{\!}$ cues$_{\!}$ and$_{\!}$ encodes$_{\!}$~the full episode history of navigation; \textbf{ii)} a \textit{target} \textit{parser} (\S\ref{sec:TP}) that actively interprets~the target specification into several embeddings; and \textbf{iii)} a \textit{multitask} \textit{planner} (\S\ref{sec:MP}) that uses the target embeddings to query encoded episodic history and leverages the returned context for action prediction.

\vspace{-3pt}
\subsection{Episodic Encoder}\label{sec:ME}
\vspace{-1pt}
At the start of each episode, \textsc{Vienna} receives a target description $g\!\in\!\{$goal image, target sound,~target class,$_{\!}$~language$_{\!}$ instruction$\}$$_{\!}$ and$_{\!}$ derives$_{\!}$ an$_{\!}$ embedding$_{\!}$ vector$_{\!}$ $\bm{g}\!\in\!\mathbb{R}^{d\!}$ (detailed$_{\!}$ in\!~\S\ref{sec:TP}).$_{\!}$ At$_{\!}$ each$_{\!}$ epoch$_{\!}$~$t$, \textsc{Vienna}$_{\!}$ has$_{\!}$ a$_{\!}$ 360$^{\circ\!}$ egocentric$_{\!}$ audiovisual$_{\!}$ perception$_{\!}$ $o_t$,$_{\!}$ \ie,$_{\!}$ RGB$_{\!\!}$ +$_{\!\!}$ depth$_{\!\!}$ +$_{\!\!}$ audio,$_{\!}$ of$_{\!}$ its$_{\!}$ surrounding.

\noindent\textbf{Intra-Modal$_{\!}$ Encoders.}$_{\!}$ A$_{\!}$ \textit{visual$_{\!}$ encoder}$_{\!\!}$ $f_{\textsc{Img}\!}$ maps$_{\!}$ perceived$_{\!}$ panoramic$_{\!}$ image$_{\!}$ {$V_t\!\in\!\mathbb{R}^{12\times224_{\!}\times_{\!}224_{\!}\times_{\!}3\!}$}
into$_{\!}$ visual$_{\!}$ features$_{\!\!}$  {$\bm{V}_{t\!}\!=\![\bm{v}_{1,t},\cdots_{\!},\bm{v}_{12,t}]\!\in\!\mathbb{R}^{12\times_{\!} d}$}, where$_{\!}$ $\bm{v}_{i,t\!}\!\in\!\mathbb{R}^{d\!}$ is$_{\!}$ the$_{\!}$ feature$_{\!}$ vector$_{\!}$ of$_{\!}$ $i$-th$_{\!}$ sub-view in$_{\!}$~$V_t$.$_{\!}$ Similarly,$_{\!}$ a$_{\!}$ \textit{depth$_{\!}$~encoder}$_{\!}$ $f_{\textsc{Dep}\!}$ and$_{\!}$ an$_{\!}$ \textit{audio$_{\!}$~encoder$_{\!}$} $f_{\textsc{Aud}\!}$ map$_{\!}$ the$_{\!}$ perceived$_{\!}$ panoramic$_{\!}$ depth image$_{\!}$ {$D_t\!\in\!\mathbb{R}^{12\times256{\!}\times_{\!}256{\!}\times_{\!}1\!}$} and$_{\!}$ spectogram$_{\!}$ tensor$_{\!}$ of$_{\!}$ binaural$_{\!}$ sound$_{\!}$ (collected$_{\!}$ over$_{\!}$ 12$_{\!}$ horizontal$_{\!}$~di- rections)$_{\!}$ {$H_t\!\in\!\mathbb{R}^{12_{\!}\times_{\!}41_{\!}\times_{\!}44_{\!}\times_{\!}2\!}$} into$_{\!}$ depth$_{\!}$ and$_{\!}$ audio$_{\!}$ features, \ie, $\bm{D}_{t\!}\!=\![\bm{d}_{1,t},\cdots_{\!},\bm{d}_{12,t}]\!\in\!\mathbb{R}^{12\times_{\!} d}$, and $\bm{A}_{t\!}\!=\![\bm{a}_{1,t},\cdots_{\!},\bm{a}_{12,t}]\!\in\!\mathbb{R}^{12\times_{\!} d}$, respectively.

\noindent\textbf{Target-Guided Cross-Modal Encoder.} With the target description vector $\bm{g}\!\in\!\mathbb{R}^{d}$, cross-attention ${f}_{\textsc{Att}\!}$ (\textit{cf}.$_{\!}$~Eq.$_{\!\!}$~\ref{equ:attn})$_{\!}$ is$_{\!}$ separately$_{\!}$ applied$_{\!}$ over$_{\!}$ $\bm{V}_{t}$,$_{\!}$ $\bm{D}_{t}$,$_{\!}$ and$_{\!}$ $\bm{H}_{t}$ to$_{\!}$ assemble$_{\!}$ target-related$_{\!}$ sensory$_{\!}$ information:
\begin{equation}\label{equ:3}
	\begin{aligned}	\tilde{\bm{v}}_t\!=\!{f}_{\textsc{Att}}(\bm{g},\bm{V}_t)\!\in\!\mathbb{R}^{d\!},~~\tilde{\bm{d}}_t\!=\!{f}_{\textsc{Att}}(\bm{g},\bm{D}_t)\!\in\!\mathbb{R}^{d\!},~~\tilde{\bm{h}}_t\!=\!{f}_{\textsc{Att}}(\bm{g},\bm{H}_t)\!\in\!\mathbb{R}^{d\!}.
	\end{aligned}
\end{equation}
Then $\tilde{\bm{v}}_t$, $\tilde{\bm{d}}_t$, and $\tilde{\bm{h}}_t$ are concatenated for attention based \textit{multisensory information integration} (MSI):
\begin{equation}
	\begin{aligned}	\bm{o}_t\!=\!{f}_{\textsc{Msi}}([\tilde{\bm{v}}_{t}, ~\tilde{\bm{d}}_{t}, ~\tilde{\bm{h}}_{t}])\!\in\!\mathbb{R}^{3_{\!}\times_{\!}d},
	\end{aligned}
    \label{equ:msi}
\end{equation}
where$_{\!}$ ${f}_{\textsc{Msi}}$$_{\!}$ is$_{\!}$ achieved$_{\!}$ by$_{\!}$ stacking$_{\!}$ two$_{\!}$ self-attention$_{\!}$ based$_{\!}$ Transformer$_{\!}$ blocks$_{\!}$ (\textit{cf}.$_{\!}$~Eq.$_{\!}$~\ref{equ:tb}).

\noindent\textbf{Episodic History Encoder.} At epoch $t$, the multimodal observation embedding $\bm{o}_{t\!}\!\in\!\mathbb{R}^{3_{\!}\times_{\!}d}$ and latest navigation action embedding $\bm{a}_{t-1\!}\!\in\!\mathbb{R}^{d\!}$, are together projected into a compact ``\textit{navigation token}'':
\begin{equation}
	\begin{aligned}	
\bm{e}_t\!=\![\bm{o}_t, \bm{a}_{t-1}]\bm{W}^e\!\in\!\mathbb{R}^{d}.
	\end{aligned}
\end{equation}
All the past navigation tokens, $\bm{e}_{1:t\!~}$, summed with corresponding epoch embedding vectors, $\bm{\mu}_{1:t}\!\in\!\mathbb{R}^{d}$, are collected into a sequence and fed into an \textit{episode history encoder} (EHE) to get contextualized history representation:
\begin{equation}
[\tilde{\bm{e}}_1,\cdots,\tilde{\bm{e}}_t]=f_{\textsc{Ehe}}([\bm{e}_{1\!}+_{\!}\bm{\mu}_1, \cdots, \bm{e}_{t\!}+_{\!}\bm{\mu}_t]),
\label{equ:ehe}
\end{equation}
where$_{\!}$ $f_{\textsc{Ehe\!}}$ is implemented as four self-attention based Transformer blocks (\textit{cf}.$_{\!}$~Eq.$_{\!}$~\ref{equ:tb}). In this way, \textsc{Vienna} is able to store and access its entire episode history of audiovisual observations and actions, leading to persistent memorization$_{\!}$ and long-term reasoning. The attended history representation $\tilde{\bm{e}}_{1:t}$ will 
\begin{wrapfigure}[6]{r}{0.62\linewidth}
	\vspace{-2mm}
	\includegraphics[width=\linewidth]{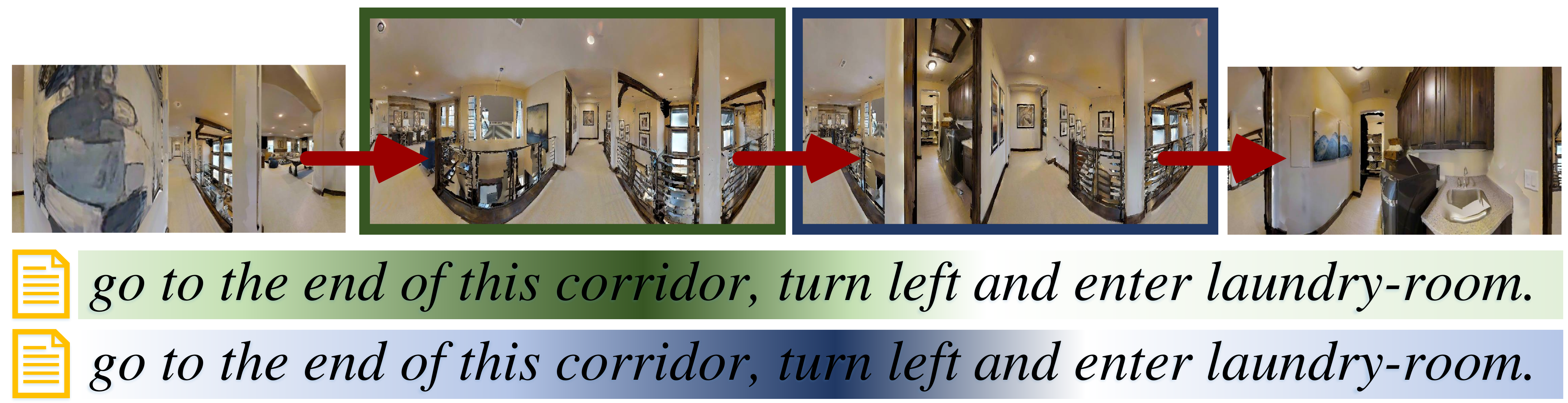}
	\caption{{$_{\!}$Attention$_{\!}$ visualization$_{\!}$} {of$_{\!}$ online$_{\!}$ target$_{\!}$ parsing$_{\!}$ (Eq.$_{\!}$~\ref{equ:q}).}}
	\label{fig:attention}
\end{wrapfigure}
serve as informative context for predicting the navigation action $a_t$ at epoch $t$ (detailed in$_{\!}$~\S\ref{sec:MP}).

\subsection{Target Parser}\label{sec:TP}

\textsc{Vienna} is equipped with a \textit{target parser} that actively interprets the target $g$ (no matter it is specified as a photo $g_{{I}}$, sound $g_{{A}}$, semantic tag $g_{{T}}$, or linguistic instruction $g_{{L}}$) as a group of target embeddings, conditioned on the progress of the navigation episode. Guided by$_{\!}$ the online$_{\!}$ created$_{\!}$ target$_{\!}$ embeddings, valuable context are selected from episodic experiences $\tilde{\bm{e}}_{1:t}$ for flexible decision-making.$_{\!}$~\textsc{Vienna} thus$_{\!}$ formulates$_{\!}$ various navigation tasks in a unified scheme, allowing to exploit cross-task knowledge.

In \textit{image-goal nav.}, a goal image {$g_{{I}\!}\in_{\!}\!\mathbb{R}^{224_{\!}\times_{\!}224_{\!}\times_{\!}3\!}$} is given and$_{\!}$ embedded$_{\!}$ as$_{\!}$
$\bm{g}_{{I}\!}\!=_{\!}\!f_{\textsc{Img}}(g_{{I}})_{\!}\!\in\!\mathbb{R}^{N_{\!I\!}\times_{\!} d}$.$_{\!}$ In$_{\!}$ \textit{audio-goal$_{\!}$ nav.}, the target is signaled by the binaural sound, \ie, {$g_{{A}\!}\!=_{\!}\!H_{t\!}\!\in_{\!}\!\mathbb{R}^{12_{\!}\times_{\!}41_{\!}\times_{\!}44_{\!}\times_{\!}2\!}$} and $\bm{g}_{{A}\!}\!=_{\!}\!\bm{H}_{t\!}\!=_{\!}\!f_{\textsc{Aud}}(H_t)_{\!}\!\in_{\!}\!\mathbb{R}^{N_{\!A\!}\times_{\!} d}$. In \textit{object-goal nav.}, the target is specified by a semantic tag $g_{{T}\!}\!\in_{\!}\!\{$$\text{table}, \text{bed}, \cdots\}$, and embedded into a class vector $\bm{g}_{{T}\!}\!\in_{\!}\!\mathbb{R}^{1_{\!}\times_{\!}d}$. In \textit{vision-language nav.}, a language-based trajectory instruction $g_{\!~\text{L}}$ is given and mapped into a sequence of word vectors $\bm{g}_{{L}\!}\!\in\!\mathbb{R}^{N_{\!L\!}\times_{\!} d}$ by a bi-LSTM.  At the start of each episode, we first build
an \textit{augmented target description embedding} $\bm{G}\!\in\!\mathbb{R}^{4N_{G\!}\times_{\!} d}$:
\begin{equation}\label{equ:g} \bm{G}\!=\!\big[\bm{g}'_{{I}}\!+\![\bm{\tau}_{{I}}]^{N_{G}\!},~\bm{g}'_{{A}}\!+\![\bm{\tau}_{{A}}]^{N_{G}\!},~\bm{g}'_{{T}}\!+\![\bm{\tau}_{{T}}]^{N_{G}\!},~\bm{g}'_{{L}}\!+\![\bm{\tau}_{{L}}]^{N_{G}}\big],
\end{equation}
where$_{\!}$ $N_{G\!}\!=\!\text{max}(N_{\!I}, N_{\!A}, 1, N_{\!L})$,$_{\!}$ $\bm{\tau}_{{I, A, T, L}\!\!}\!\in\!\mathbb{R}^{d\!}$ are$_{\!}$ learnable task embedding vectors, and~$[\!\!~\cdot~\!\!]^{N_{G}\!}$ copies$_{\!}$ its$_{\!}$ input$_{\!}$ $N_{G\!}$ times.$_{\!}$ Assuming$_{\!}$ \textsc{Vienna}$_{\!}$ is$_{\!}$ in$_{\!}$ an$_{\!}$ \textit{image-goal$_{\!}$ nav.}$_{\!}$~episode,$_{\!}$ we$_{\!}$ have$_{\!}$ $\bm{g}_{{I}\!}\!=_{\!}\!f_{\textsc{Img}}(g_{{I}})\!\in\!\mathbb{R}^{N_{\!I\!}\times_{\!} d}$, and $\bm{g}_{{A}\!}\!=_{\!}\![\bm{0}]^{N_{\!A\!}\times_{\!} d}$, $\bm{g}_{{T}\!}\!=_{\!}\![\bm{0}]^{1_{\!}\times_{\!} d}$, $\bm{g}_{{L}\!}\!=_{\!}\![\bm{0}]^{N_{\!L\!}\times_{\!} d}$. We pad $\bm{g}_{{I, A, T, L}\!}$ to a unified length~$N_{G}$, by replication, so as to get $\bm{g}'_{{I, A, T, L}\!}$ and make them contribute equally to $\bm{G}$. We collect all the task-type embeddings $\bm{\tau}_{I, A, T, L\!}$ and current target description~$g_{\!}\in_{\!}\{g_{{I}}, g_{{A}}, g_{{T}}, g_{{L}}\}$ into $\bm{G}$. In~\S\ref{sec:MP}, we~will show this strategy is essential for making use of cross-task knowledge. The target description vector $\bm{g}\!\in\!\mathbb{R}^{d}$ used in Eq.~\ref{equ:3} is given as: $\bm{g}\!=\!f_{\textsc{Avg}}(\bm{G})$, where $f_{\textsc{Avg}\!}$ stands for the average pooling operation.

At epoch $t$, the target parser comprehends the augmented target description embedding $\bm{G}$ as a set of $N$ compact embeddings on-the-fly, conditioned on its episodic, contextualized history encoding $\tilde{\bm{e}}_{1:t}$:
\begin{equation}\label{equ:q}	
\bm{Q}_t\!=\![\bm{q}^1_t,\cdots,\bm{q}^N_t]\!=\!f_{\textsc{Mha}}(f_{\textsc{Avg}}(\tilde{\bm{e}}_{1:t}), \bm{G})\!\in\!\mathbb{R}^{N\times d},
\end{equation}
where$_{\!}$ $f_{\textsc{Mha}\!}$ is$_{\!}$ a$_{\!}$ $N$-head$_{\!}$ attention$_{\!}$ layer$_{\!}$ (\textit{cf}.$_{\!}$~Eq.$_{\!}$~\ref{equ:tb}),$_{\!}$ \ie, explain  $\bm{G}$ in different ways, with consideration of current episodic navigation progress. Each of the target embedding vectors $\bm{q}_t$ can be viewed as a specific, time-varying goal, used to guide action selection $a_t$ at epoch $t$. As shown in Fig.$_{\!}$~\ref{fig:attention}, given~a navigation instruction ``\textit{go to the end of this corridor, turn left and $\cdots$}'', the agent focuses more on ``\textit{go to the end of this corridor}'' at the start of the navigation episode.  After reaching the end of the corridor, the agent shifts its attention to  ``\textit{turn left}''. Here a collection of $N$ target embeddings $\bm{q}^{1:N\!\!}$ are generated at each epoch $t$, allowing the agent to capture different aspects of target-related information and making the time-varying goal well-planned. For instance, there may exist several essential landmarks in a goal image, or multiple discriminative audio clips in target-emitted sound; during navigation, the agent should be able to pay attention to all these informative clues simultaneously.

\subsection{Multitask Planner}\label{sec:MP}
At epoch $t$, a multitask planner (MTP) uses the diversified target embeddings $\bm{Q}_{t\!}\!=_{\!}\![\bm{q}^1_t,\cdots_{\!},\bm{q}^N_t]$ to query the episodic history $\tilde{\bm{e}}_{1:t}$~\!:
\begin{equation}\label{eq:MTP}	
~~~~\bm{C}_t\!=\!f_{\textsc{Mtp}}([\bm{q}^1_t,\cdots_{\!},\bm{q}^N_t],~~ [\tilde{\bm{e}}_{1},\cdots_{\!},\tilde{\bm{e}}_{t}]),
\end{equation}
where $f_{\textsc{Mtp}}$ is achieved by a four-layer Transformer decoder; the first two layers are shared among the four navigation tasks for capturing task-shared policies, while the last two layers are private for each task for task-specific policy learning. We empirically find such \textit{shared trunk} based MTP design yields better performance than learning task-specific policies individually or just training one single ``universal'' policy (\textit{cf}.$_{\!}$~\S\ref{sec:ABS}). 

The decision-making is conditioned on the retrieved context $\bm{C}_{t\!}\!\in_{\!}\!\mathbb{R}^{N_{\!}\times_{\!}d\!}$, and the presentations of current multi-modal observations (\textit{cf}.$_{\!}$~\S\ref{sec:dd}), including $\bm{V}_{t\!}\!=\![\bm{v}_{1,t},\cdots_{\!},\bm{v}_{12,t}]\!\in\!\mathbb{R}^{12\times_{\!} d}$, $\bm{D}_{t\!}\!=\![\bm{d}_{1,t},\cdots_{\!},\bm{d}_{12,t}]\!\in\!\mathbb{R}^{12\times_{\!} d}$, and $\bm{A}_{t\!}\!=\![\bm{a}_{1,t},\cdots_{\!},\bm{a}_{12,t}]\!\in\!\mathbb{R}^{12\times_{\!} d}$. Specifically, at epoch $t$, \textsc{Vienna} makes navigate decision by choosing between the 12 current sub-views, as well as an extra \texttt{STOP} action. Given 12 subview action embeddings $\{\bm{b}_{i,t}\!\in\!\mathbb{R}^{3\times_{\!} d}\}^{12}_{i=1}$, \ie, $\bm{o}_{i,t}\!=\![\bm{v}_{i,t}, \bm{d}_{i,t}, \bm{h}_{i,t}]$ as well as a \texttt{STOP} action embedding, \ie, $\bm{b}_{13,t}\!=\!\vec{0}$, represented as an all-zero vector, \textsc{Vienna}  predicts a probability distribution $\bm{p}_{t\!}\!=\![p_{1,t},\cdots_{\!},p_{13,t}]$:
\begin{equation}
    {p}_{i,t}=\text{softmax}_i(f_{\textsc{Avg}}(\bm{C}_t)\bm{W}^p\bm{b}_{i,t}) \in [0,1], ~~~~~~~~~\text{where}~~i\in\{1,\cdots,13\}.
\end{equation}
As the task embeddings $\bm{\tau}_{I, A, T, L\!}$ are encoded into $\bm{Q}_{t}$, which is used to find supportive cues from episodic observations $\tilde{\bm{e}}_{1:t\!}$ for long-term reasoning and decision-making, $\bm{\tau}_{I, A, T, L\!}$ are essentially trained as task-wise context -- they are sensitive to task-related cues. Thus collecting $\bm{\tau}_{I, A, T, L\!}$ into $\bm{G}$$_{\!}$ (\textit{cf}.$_{\!}$~Eq.$_{\!}$~\ref{equ:g}) enables a clever use of cross-task knowledge. For instance, during \textit{image-goal nav.}, $\bm{\tau}_{{A}\!}$ can help the agent notice some informative audio signals, $\bm{\tau}_{{T}\!}$ can alert the agent to visually essential semantics, while $\bm{\tau}_{{I}\!}$ can be activated by crucial landmarks. Related experiments can be found in \S\ref{sec:ABS}.

\subsection{MTRL based Multitask Navigation Training}\label{sec:MDPPO}

\noindent\textbf{Reward Design.$_{\!}$} With$_{\!}$ standardized$_{\!}$ \textbf{\texttt{VXN}}$_{\!}$ environments,$_{\!\!}$ \textsc{Vienna}$_{\!}$ adopts$_{\!}$ a$_{\!}$ same$_{\!}$ reward~function for the four$_{\!}$ navigation$_{\!}$ tasks,$_{\!}$ \ie,$_{\!}$ $R_{1\!}\!=_{\!}\!\cdots_{\!}\!=_{\!}\!R_4$.$_{\!}$ Concretely,$_{\!}$~$R_{1:4\!}$~has four terms, \ie, a sparse success reward $r_{\text{success}}$, a progress reward $r_{\text{progress}}$, a slack reward $r_{\text{slack}}$, and an exploration reward $r_{\text{explore}}$. $r_{\text{success}\!}\!=\!2.5$ is$_{\!}$ only$_{\!}$ received$_{\!}$ at$_{\!}$ the$_{\!}$ end$_{\!}$ of$_{\!}$ a$_{\!}$ successful$_{\!}$ episode.$_{\!}$ $r_{\text{progress}\!}\!=_{\!}\!-\Delta_{\text{geo\_dist}}$ offers$_{\!}$ dense$_{\!}$ signals indicating the progress that an action contributes: $\Delta_{\text{geo\_dist}}$ gives the change in geodesic~distance to the goal position by performing the action. $r_{\text{slack}\!}\!=_{\!}\!-10^{-3\!}$, received at each~epoch, penalizes redundant actions.$_{\!}$ $r_{\text{explore}\!}$~\cite{ye2021auxiliary}$_{\!}$ divides$_{\!}$ each$_{\!}$ environment$_{\!}$ into$_{\!}$ a$_{\!}$ voxel$_{\!}$ grid$_{\!}$ with$_{\!}$ $2.5\!~\text{m}\!\times\!2.5\!~\text{m}\!\times\!2.5\!~\text{m}$$_{\!}$~voxels$_{\!}$~and rewards$_{\!}$ the$_{\!}$ agent$_{\!}$ for$_{\!}$ visiting$_{\!}$ each$_{\!}$ voxel.$_{\!}$  $r_{\text{explore}\!}$ is$_{\!}$ defined$_{\!}$ as$_{\!}$ $0.25\eta$,$_{\!}$ where$_{\!}$ $\eta\!=\!\delta^{t\!}/\nu_{\!}$ is a coefficient that decays as episode epoch $t$ and visited voxel number $\nu$ increase, and $\delta$ is a decay constant of $0.995$.

\noindent\textbf{Multitask$_{\!}$ Distributed$_{\!}$ Proximal$_{\!}$ Policy$_{\!}$ Optimization.}$_{\!}$ We present$_{\!}$ a$_{\!}$ multitask$_{\!}$ distributed$_{\!}$ proximal$_{\!}$ policy$_{\!}$ optimization (MDPPO) algorithm, which utilizes the power of parallel processing to train MTRL agents in our continuous and large-scale environments. MDPPO is built upon (DPPO)$_{\!}$~\cite{heess2017emergence}, a distributed version of proximal policy optimization (PPO)$_{\!}$~\cite{schulman2017proximal} that bounds
parameter updates to a trust region to ensure stability, and  distributes the computation over many parallel instances of agent and environment. Similarly, MDPPO has a server-client structure: each~client worker$_{\!}$ has$_{\!}$ several$_{\!}$ agent$_{\!}$ copies that collect experiences from \textbf{\texttt{VXN}}$_{\!}$ environments,$_{\!}$ compute$_{\!}$ and$_{\!}$ send$_{\!}$ PPO's$_{\!}$ gradients$_{\!}$ to$_{\!}$ the server; the server worker averages the received gradients, updates the agent, and synchronizes the$_{\!}$ updated$_{\!}$ weights$_{\!}$ with$_{\!}$ the clients. For balanced multitask learning, \ie, training data in \textbf{\texttt{VXN}}  are biased between \textit{vision-language nav.} and other navigation tasks: $10.8\text{K}$ \textit{vs} $2.0_{\!}\!\sim_{\!}\!5.0\text{M}$ episodes (\textit{cf}.$_{\!}$~Table$_{\!}$~\ref{tab:dataset}), each client worker is required to build four agent copies corresponding to the four \textbf{\texttt{VXN}} tasks.

\subsection{Implementation Detail}\label{sec:ID}

\noindent\textbf{Network$_{\!}$ Architecture.}$_{\!}$ The$_{\!}$ \textit{visual$_{\!}$ encoder}$_{\!}$ $f_{\textsc{Img}}$$_{\!}$ is$_{\!}$ made$_{\!}$ as an ImageNet$_{\!}$~\cite{ILSVRC15}-pretrained ResNet50 $_{\!}$~\cite{he2016deep}.  The CNN~features are fed into a linear layer for dimension compression and flattened into a feature sequence. Similarly, the \textit{depth encoder} $f_{\textsc{Dep}}$$_{\!}$ is a modified ResNet50 CNN. The \textit{audio encoder} $f_{\textsc{Aud}}$, following$_{\!}$~\cite{chen2020soundspaces}, is a CNN of conv $8\!\times\!8$, conv $4_{\!}\times_{\!}4$,$_{\!}$ conv$_{\!}$ $3_{\!}\times_{\!}3$ and$_{\!}$ a$_{\!}$ linear$_{\!}$ layer,$_{\!}$ interleaved$_{\!}$ with$_{\!}$ ReLU.$_{\!}$ All the sensory features are combined with orientation embeddings. For the \textit{epoch embedding} $\bm{\mu}$, we use sinusoidal encoding. For the \textit{target parser}, $N\!=\!5$ target embeddings are generated at each epoch $t$. We set other hyper-parameters as:$_{\!}$ $d_{\!}\!=_{\!}\!512$, $N_{\!I\!}\!=_{\!}\!16$, $N_{\!L\!}\!=_{\!}\!120$, $N_{G\!}\!=_{\!}\!120$.

\noindent\textbf{Training and Test.} \textsc{Vienna} is trained on 32 RTX 2080 GPUs {for $180$ M frames}, costing $4,608$ GPU hours. As in \cite{krantz2020navgraph}, we select the checkpoint for evaluation with the best SR on \texttt{val} \texttt{unseen}. For MDPPO, we use four client workers and set the discounted factor $\gamma$ as $0.99$. We use AdamW$_{\!}$~\cite{loshchilov2018fixing} optimizer with a learning rate of $2.5\!\times\!10^{-4}$. Casual attention$_{\!}$~\cite{vaswani2017attention} is adopted to prevent the prediction at epoch $t$ from the influence of future tokens after $t$. Once trained, a single instance of \textsc{Vienna} can conduct the four navigation tasks. As normal, greedy prediction is adopted for action selection.

\section{Experiment}

In \S\ref{sec:pc}, we first report comparison results for the four \textbf{\texttt{VXN}} tasks. In \S\ref{sec:ABS}, we conduct diagnostic studies to examine the efficacy of our core model design. More results are put in the \textbf{\textit{supplementary}}.

\noindent\textbf{Baseline.$_{\!}$} We$_{\!}$ test$_{\!}$ several$_{\!}$ open-source$_{\!}$ task-specific$_{\!}$ navigation$_{\!}$ methods$_{\!}$~\cite{zhu2017target,chen2020soundspaces,chaplot2020object,krantz2020navgraph}. Note that~\cite{zhu2017target,chen2020soundspaces,chaplot2020object} are re-trained on \textbf{\texttt{VXN}}, since they use different training data~\cite{zhu2017target}, world representation~\cite{chen2020soundspaces} (discrete \textit{vs} continuous), or object categories~\cite{chaplot2020object} (6 \textit{vs} 21). For~\cite{krantz2020navgraph}, we use its check-point but the success criteria are different (3 m \textit{vs} 1 m). Thus their scores on \textbf{\texttt{VXN}} are different from the original ones.  In addition, we$_{\!}$ consider$_{\!}$ a$_{\!}$ \texttt{Seq2Seq}$_{\!}$ agent,$_{\!}$ which$_{\!}$ also$_{\!}$ serves$_{\!}$ as$_{\!}$ a$_{\!}$ standard$_{\!}$ baseline$_{\!}$ in$_{\!}$~\cite{anderson2018vision,krantz2020navgraph}:$_{\!}$ an$_{\!}$ LSTM$_{\!}$ planner$_{\!}$~encodes the episode history
and predicts navigation actions in a sequential menner. For all~the four tasks, we provide the performance of both the single-task and multitask versions of our \textsc{Vienna} and \texttt{Seq2Seq}. Further, \texttt{Random} \texttt{policy}, \ie,  choosing actions randomly, is included.

\noindent\textbf{Metric.} Four widely-used metrics are adopted for evaluation: i) \textit{Success Rate} (\textbf{SR}); ii) \textit{Navigation Error} (NE); iii) \textit{Oracle success Rate} (OR); and iv) \textit{Success rate weighted by Path Length} (SPL)~\cite{anderson2018evaluation}.

\subsection{Performance Benchmarking}\label{sec:pc}
Table$_{\!}$~\ref{table:qc} reports the comparison results on the four \textbf{\texttt{VXN}} tasks. Some key conclusions are list below:
\begin{itemize}[leftmargin=*]
	\setlength{\itemsep}{0pt}
	\setlength{\parsep}{-0pt}
	\setlength{\parskip}{-0pt}
	\setlength{\leftmargin}{-6pt}

     \item \textsc{Vienna} obtains impressive results, under \texttt{val} \texttt{seen} and \texttt{unseen} sets,  across all the tasks and~evaluation metrics. This proves the versatility of \textsc{Vienna} and the power of our parse-and-query~regime.
    \item \textsc{Vienna} consistently outperforms \texttt{Seq2Seq}, no matter they are trained on single tasks individually or multiple tasks jointly. {Compared with other task-specific competitors~\cite{zhu2017target,chen2020soundspaces,chaplot2020object,krantz2020navgraph}, VIENNA gains comparable results on \textit{audio-goal nav}. and \textit{object-goal nav}., and performs better on \textit{image-goal nav}. and \textit{vision-language nav}. tasks.} These results verify the effectiveness of our model design.
    \item When considering the performance gain from the single-task setting to multitask, \textsc{Vienna} {yields more promising results,} compared with \texttt{Seq2Seq}. For example, in Table$_{\!}$~\ref{table:IGN}, {\textsc{Vienna}$_{\text{MT}}$ outperforms \textsc{Vienna}$_{\text{ST}}$ by 2.2\% SR and 1.7\% SR,} on \texttt{val}~\texttt{seen} and \texttt{unseen}, respectively; however, in the same condition, \texttt{Seq2Seq}$_{\text{MT}}$ only provides 0.7\% and 0.9\% SR gains over \texttt{Seq2Seq}$_{\text{ST}}$. These results demonstrates that \textsc{Vienna} can make a better use of cross-task knowledge.
    \item When considering the performance gap between seen and unseen environments, \textsc{Vienna}$_{\text{MT}\!}$ is more favored than its single-task counterpart, \textsc{Vienna}$_{\text{ST}\!}$. For instance, in Table$_{\!}$~\ref{table:OGN}, \textsc{Vienna}$_{\text{ST}\!}$ suffers from relatively large performance drop, \ie, {33.2\%$\rightarrow$18.5\%$_{\!}$} SR;$_{\!}$ however,$_{\!\!}$ \textsc{Vienna}$_{\text{MT}\!}$ shows$_{\!}$ reduced$_{\!}$ degradation, \ie, {33.3\%$\rightarrow$19.4\% SR}, in unseen environments. {This indicates that investigating inter-task relatedness may help to strengthen the generalizability of navigation agents.}
      \item The above results are particularly impressive considering the advantage of \textsc{Vienna} in efficient
         parameter utilization, \ie, \textsc{Vienna}$_{\text{MT}\!}$ (31 M) \textit{vs} \textsc{Vienna}$_{\text{ST}\!\!}\times_{\!}4$  (101 M) \textit{vs} \texttt{Seq2Seq}$_{\text{MT}}$  (27 M) \textit{vs} \texttt{Seq2Seq}$_{\text{ST}\!\!}\times_{\!}4$  (93 M) \textit{vs} \cite{zhu2017target} +\cite{chen2020soundspaces} +\cite{chaplot2020object} +\cite{krantz2020navgraph} (165M = 40 M + 45 M + 38 M + 42 M).
\end{itemize}

\begin{figure}
    \centering
    \subfloat[\small{\textit{image-goal nav.} }]{
        \includegraphics[width=0.245\textwidth]{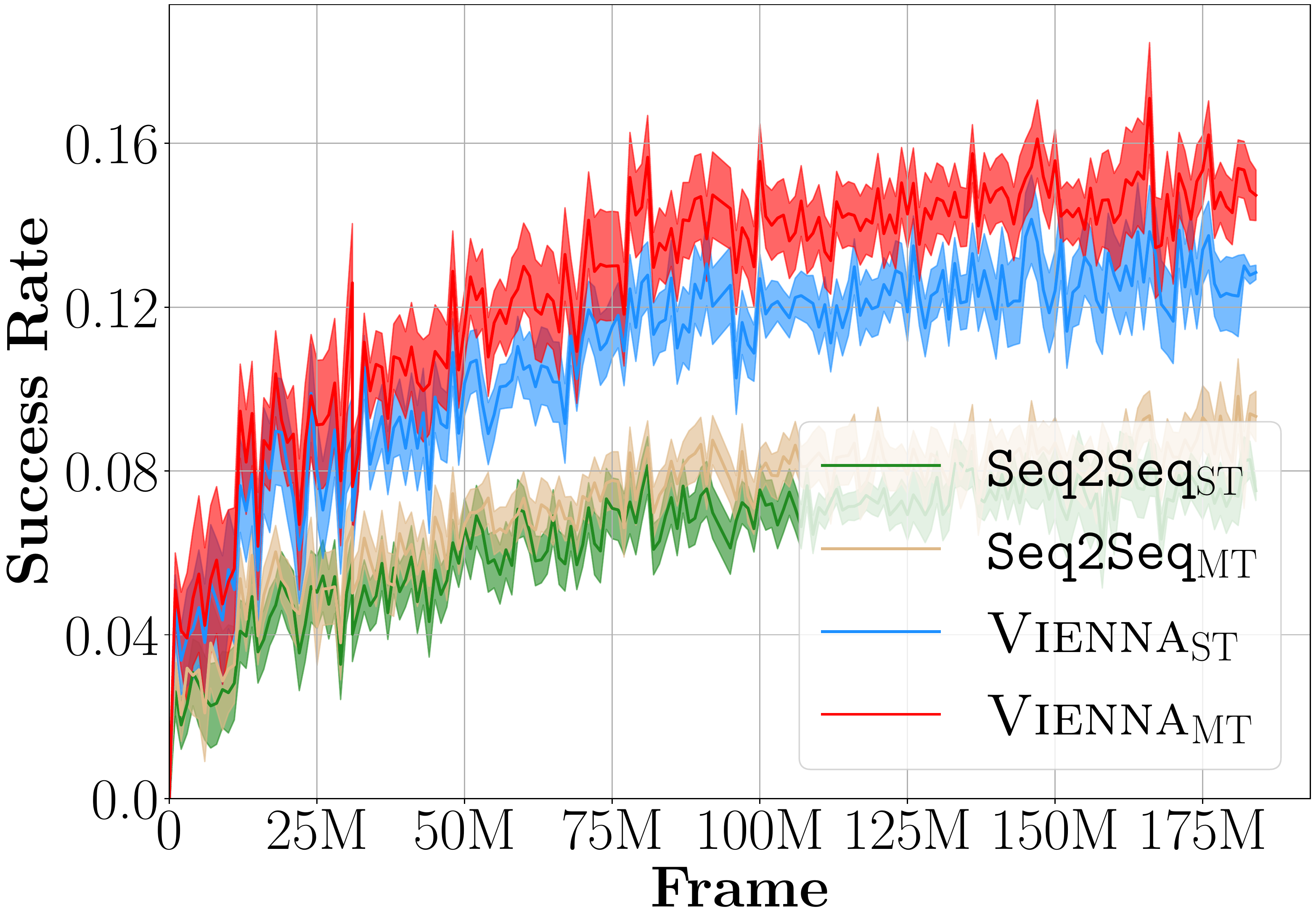}
    }
    \subfloat[\small{\textit{audio-goal nav.} }]{
        \includegraphics[width=0.245\textwidth]{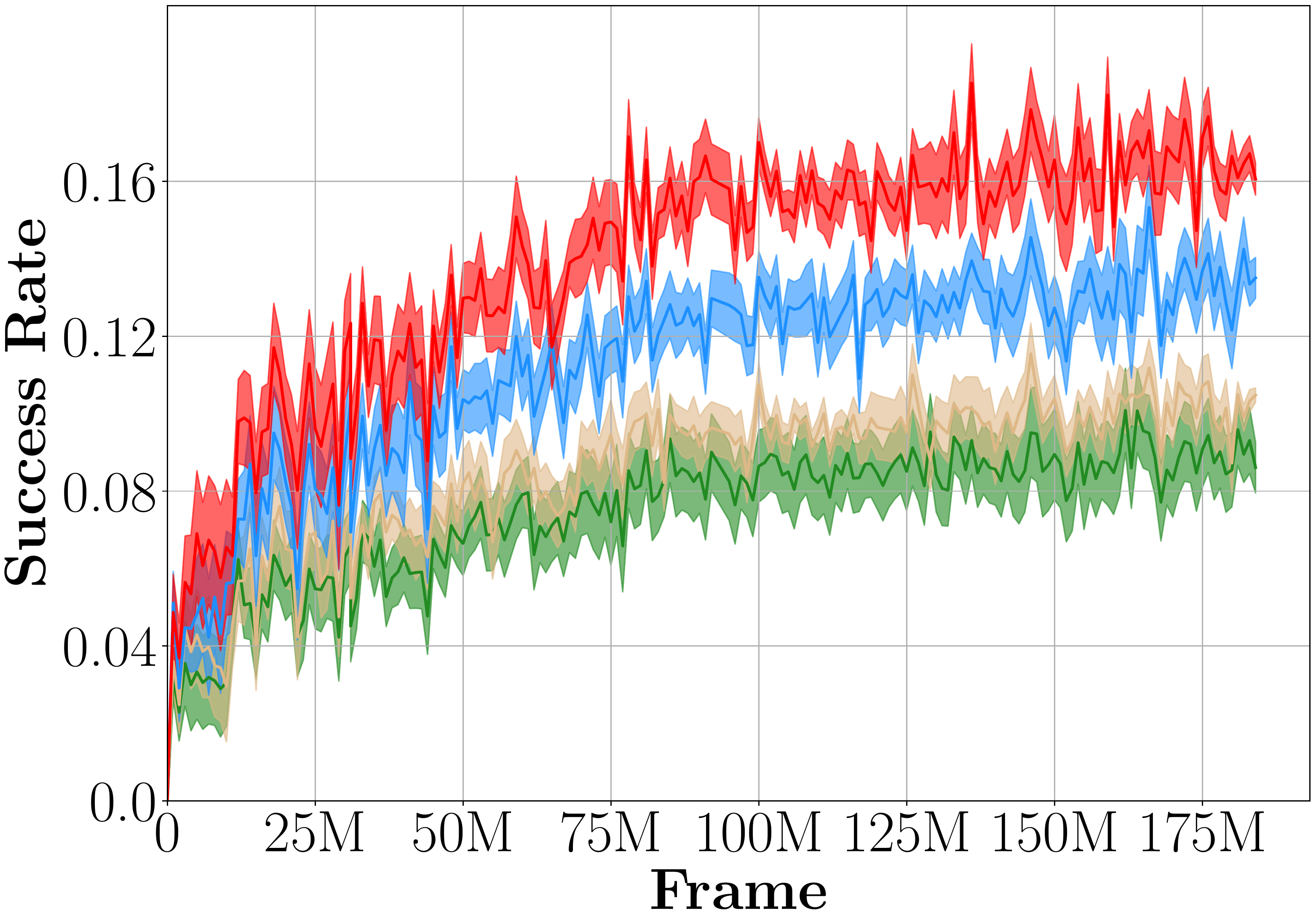}
    }
    \subfloat[\small{\textit{object-goal nav.} }]{
        \includegraphics[width=0.245\textwidth]{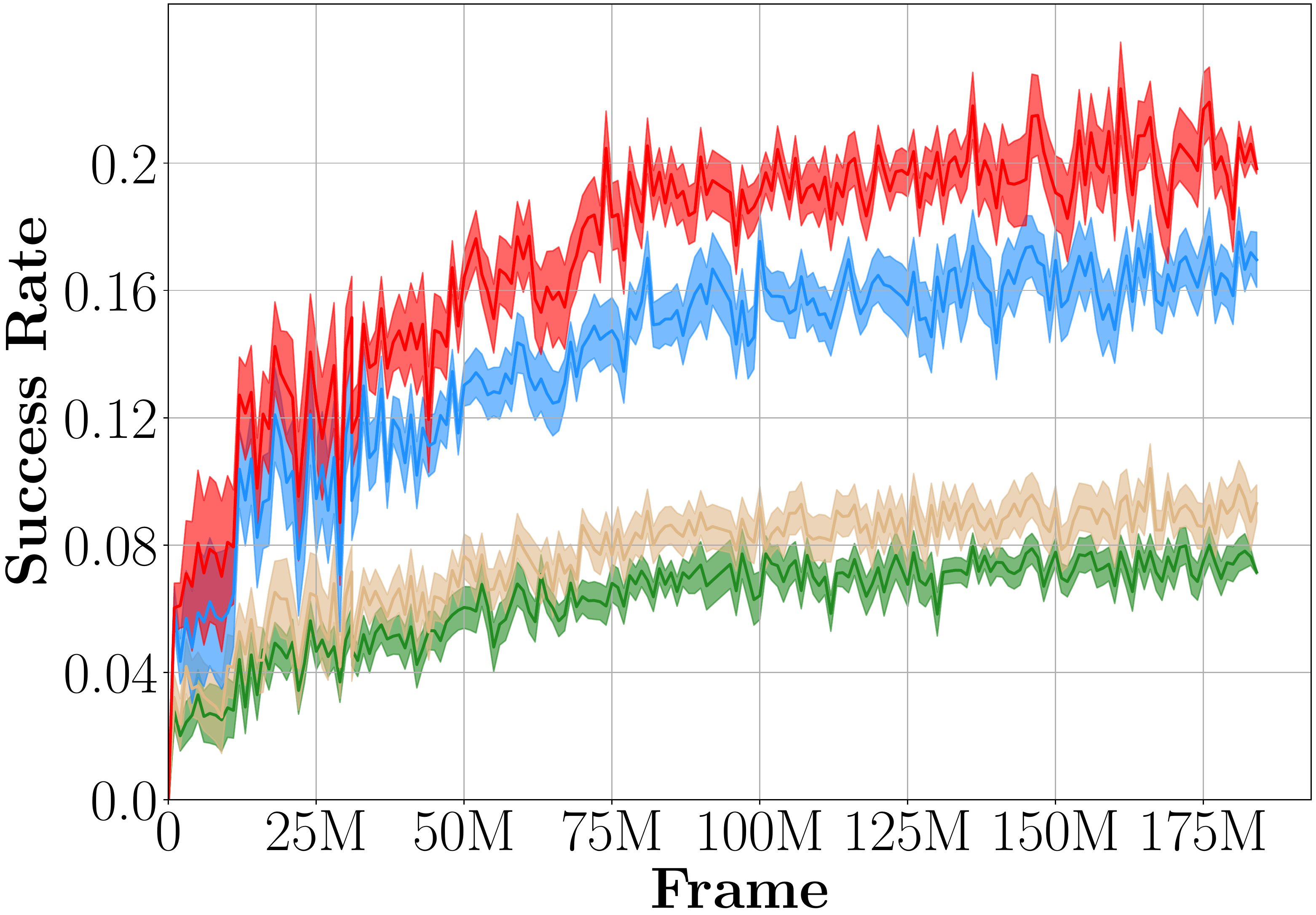}
    }
    \subfloat[\small{\textit{vision-language nav.} }]{
        \includegraphics[width=0.245\textwidth]{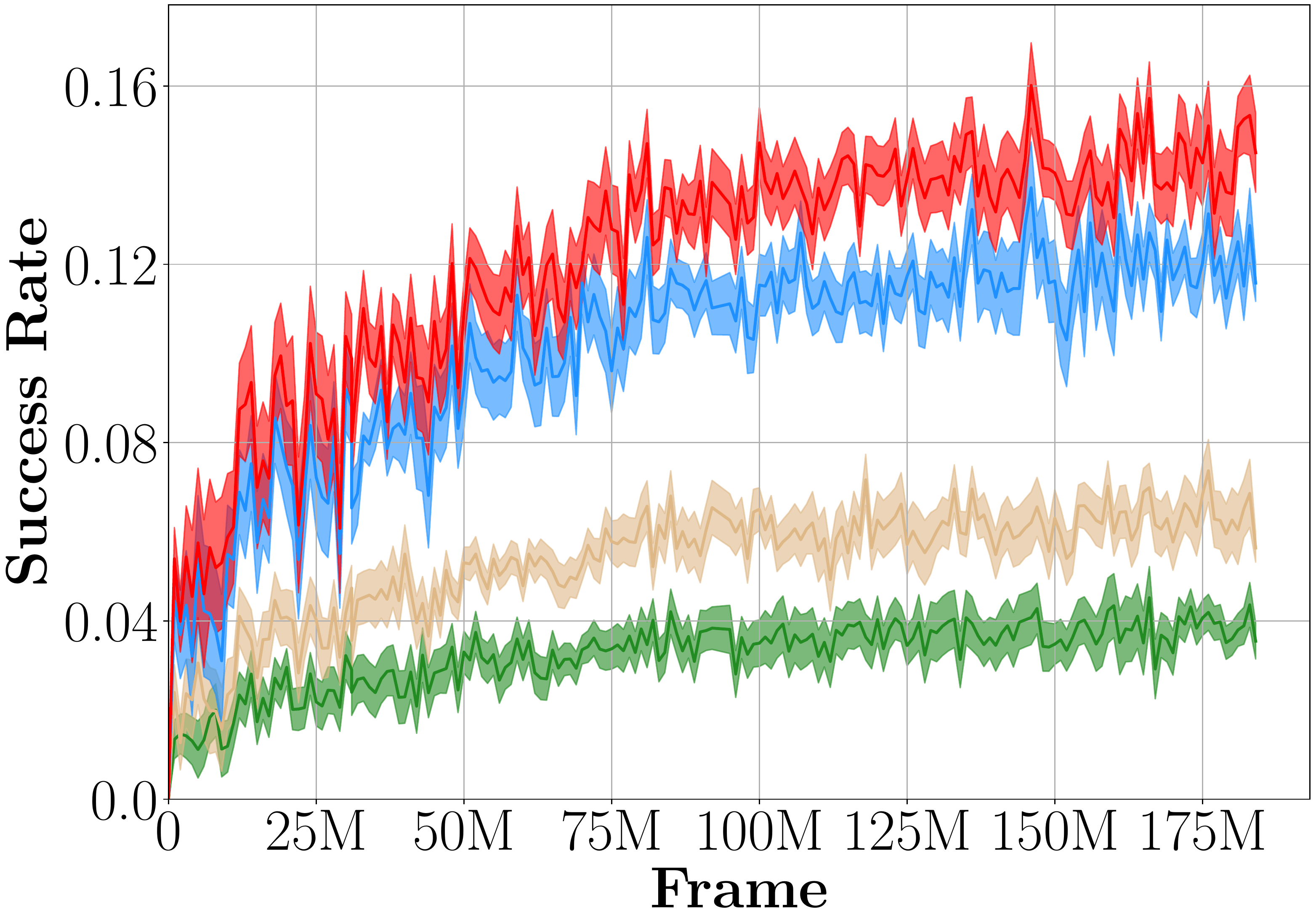}
    }
    \caption{$_{\!\!}$Training$_{\!}$ curves$_{\!}$ of$_{\!}$ \textsc{Vienna}$_{\!}$ agents$_{\!}$ compared$_{\!}$ to$_{\!}$ \texttt{Seq2Seq}$_{\!}$ agents$_{\!}$ on$_{\!}$ the$_{\!}$ four$_{\!}$ \texttt{VXN}$_{\!}$ tasks$_{\!}$ (\S\ref{sec:pc}).}
    \label{fig:curve}
\end{figure}

{Fig.$_{\!}$~\ref{fig:curve} plots the training curves of \textsc{Vienna}$_{\text{ST/MT}}$ compared to \texttt{Seq2Seq}$_{\text{ST/MT}}$ for the four \texttt{VXN} tasks in \texttt{unseen} envs. Aligning with the results in Table~\ref{table:qc}, \textsc{Vienna} outperforms \texttt{Seq2Seq}, and benefits more from multiple task learning. This shows that \textsc{Vienna} makes a better use of cross-task knowledge.}

\begin{table*}[t]
        \caption{Quantitative comparison results (\S\ref{sec:pc}) on \textbf{\texttt{VXN}} dataset {\footnotesize(\text{ST}: Single-task; \text{MT}: Multitask)}.}\label{table:qc}\vspace{-2mm}
        \hspace{-0.7em}
        \subfloat[\small{\textit{image-goal nav.} (IGN)} \label{table:IGN}]{
            \resizebox{0.51\textwidth}{!}{
            \tablestyle{4pt}{1.1}
            \begin{tabular}{|c|cccc|cccc|}
            \hline
            ~ &  \multicolumn{4}{c|}{\texttt{val} \texttt{seen}} & \multicolumn{4}{c|}{\texttt{val} \texttt{unseen}} \\
            \cline{2-9}
            \multirow{-2}{*}{Models} &\textbf{SR}$^\uparrow$ &NE$^\downarrow$  &OR$^\uparrow$  &SPL$^{\!\uparrow}$ &\textbf{SR}$^\uparrow$ &NE$^\downarrow$  &OR$^\uparrow$  &SPL$^{\!\uparrow}$\\
            \hline
            \texttt{Random}& 1.2& 14.20&1.9&1.2 & 1.4 & 14.14 & 2.2 & 1.4\\
            \texttt{Seq2Seq}$_{\text{ST}}$& 15.1&10.44& 19.1& 12.6& 9.3 &12.02 & 13.9 & 7.4\\
            \texttt{Seq2Seq}$_{\text{MT}}$& 15.8 & 10.21 & 21.3 &13.0 & 10.2& 10.22 & 15.4&8.5\\
            Zhu$_{\!}$~\etal$_{\!}$~\cite{zhu2017target}& 17.7 & 9.67 & 22.0 & 13.1 & 12.0&10.19 &16.6  & 8.9\\
            \arrayrulecolor{gray}\cdashline{1-9}[1pt/1pt]
            \textsc{Vienna}$_{\text{ST}}$& \second{19.9} & \second{9.52} & \second{23.2} & \second{13.4}& \second{12.6} & \second{9.83}  & \second{17.1}& \second{9.5}\\
            \textbf{\textsc{Vienna}}$_{\text{MT}}$& \best{22.1} & \best{9.43}& \best{24.2}& \best{14.1}& \best{14.3} & \best{9.66}& \best{18.5}& \best{11.1}\\
            \arrayrulecolor{black}\hline
            \end{tabular}
            }
            }
        \hspace{-0.5em}
        \subfloat[\small{\textit{audio-goal nav.} (AGN)}\label{table:AGN}]{%
        \resizebox{0.51\textwidth}{!}{
            \tablestyle{4pt}{1.13}
            \begin{tabular}{|c|cccc|cccc|}
            \hline
            ~ &  \multicolumn{4}{c|}{\texttt{val} \texttt{seen}} & \multicolumn{4}{c|}{\texttt{val} \texttt{unseen}} \\
            \cline{2-9}
            \multirow{-2}{*}{Models} &\textbf{SR}$^\uparrow$ &NE$^\downarrow$  &OR$^\uparrow$  &SPL$^{\!\uparrow}$ &\textbf{SR}$^\uparrow$ &NE$^\downarrow$  &OR$^\uparrow$  &SPL$^{\!\uparrow}$\\
            \hline
            \texttt{Random}& 0.0 & 17.13 &0.0&0.0& 0.0 & 16.84 &0.0 & 0.0\\
            \texttt{Seq2Seq}$_{\text{ST}}$& 17.4& 10.11& 19.0& 15.8& 11.0& 10.83&13.3& 8.8\\
            \texttt{Seq2Seq}$_{\text{MT}}$& 18.1&9.69& 20.3& 16.0& 11.8 & 10.76 & 14.1 &9.3\\
            Chen~\etal~\cite{chen2020soundspaces}& 20.1&8.84 & 21.5& 17.1& 13.1 & 9.26 & 15.7 & 10.4\\
            \arrayrulecolor{gray}\cdashline{1-9}[1pt/1pt]
            \textsc{Vienna}$_{\text{ST}}$& \second{22.4} & \second{8.76} & \second{22.4} & \second{17.3} & \second{14.3} & \second{9.22} & \second{16.5} & \second{10.6}\\
            \textbf{\textsc{Vienna}}$_{\text{MT}}$& \best{25.3} & \best{8.61} & \best{23.9} & \best{17.8}& \best{18.7} & \best{8.93} & \best{17.9}& \best{12.5}\\
            \arrayrulecolor{black}\hline
            \end{tabular}
    }}\vfill
    \vspace{-.11in}
        \hspace{-0.7em}
        \subfloat[\small{\textit{object-goal nav.} (OGN)}\label{table:OGN}]{
            \tablestyle{4pt}{1.1}
            \resizebox{0.505\textwidth}{!}{
            \begin{tabular}{|c|cccc|cccc|}
            \hline
            ~ &  \multicolumn{4}{c|}{\texttt{val} \texttt{seen}} & \multicolumn{4}{c|}{\texttt{val} \texttt{unseen}} \\
            \cline{2-9}
            \multirow{-2}{*}{Models} &\textbf{SR}$^\uparrow$ &NE$^\downarrow$  &OR$^\uparrow$  &SPL$^{\!\uparrow}$ &\textbf{SR}$^\uparrow$ &NE$^\downarrow$  &OR$^\uparrow$  &SPL$^{\!\uparrow}$\\
            \hline
            \texttt{Random}& 0.8 & 7.67 & 1.0 & 0.8 & 2.0 & 7.56 & 2.1 & 1.7\\
            \texttt{Seq2Seq}$_{\text{ST}}$& 26.7 &6.61 &33.3& 14.4 & 8.9 & 7.31  & 11.1 & 4.4\\
            \texttt{Seq2Seq}$_{\text{MT}}$& 28.7& 6.45&35.0&15.8& 10.8 &7.13& 14.0& 4.8\\
            Chaplot \etal\!~\cite{chaplot2020object}& 31.3 & 6.15 & 35.2& 16.7&  17.6& 7.08 &  21.3  & 7.5 \\
            \arrayrulecolor{gray}\cdashline{1-9}[1pt/1pt]
            \textsc{Vienna}$_{\text{ST}}$& \second{33.2} &\second{6.11}&\second{36.4}& \second{17.1}& \second{18.5} & \second{6.95} &\second{22.1}&\second{8.1}\\
            \textbf{\textsc{Vienna}}$_{\text{MT}}$& \best{33.3}&\best{5.92} & \best{37.8}& \best{17.7} & \best{19.4} &\best{6.77} &\best{25.1}&\best{10.7}\\
            \arrayrulecolor{black}\hline
            \end{tabular}
            }}
                \hspace{-0.2em}
        \subfloat[\small{\textit{vision-language nav.} (VLN)}\label{table:VLN}]{%
               \resizebox{0.51\textwidth}{!}{
            \tablestyle{4pt}{1.12}
            \begin{tabular}{|c|cccc|cccc|}
            \hline
            ~ &  \multicolumn{4}{c|}{\texttt{val} \texttt{seen}} & \multicolumn{4}{c|}{\texttt{val} \texttt{unseen}} \\
            \cline{2-9}
            \multirow{-2}{*}{Models} &\textbf{SR}$^\uparrow$ &NE$^\downarrow$ &OR$^\uparrow$  &SPL$^\uparrow$ &\textbf{SR}$^\uparrow$ &NE$^\downarrow$ &OR$^\uparrow$  &SPL$^{\!\uparrow}$\\
            \hline
            \texttt{Random}& 0.0 &8.89 &0.0 & 0.0& 0.0& 8.92 & 0.0 & 0.0\\
            \texttt{Seq2Seq}$_{\text{ST}}$& 13.2 &7.54 & 17.7 & 12.1 & 5.2&8.49 & 9.7& 4.6\\
            \texttt{Seq2Seq}$_{\text{MT}}$& 17.6 & 7.29  & 22.8 &15.4& 7.6&8.21 & 13.4& 6.5\\
            Krantz$_{\!}$~\etal$_{\!}$~\cite{krantz2020navgraph}& 23.7& 7.22 & 25.9& 21.2& 11.0 & 7.60 & 16.2 & 10.2\\
            \arrayrulecolor{gray}\cdashline{1-9}[1pt/1pt]
            \textsc{Vienna}$_{\text{ST}}$& \second{23.9} & \second{7.16}& \second{26.1}& \second{22.2}& \second{14.3}& \second{7.35} & \second{18.5} &\second{12.5}\\
            \textbf{\textsc{Vienna}}$_{\text{MT}}$& \best{26.5} & \best{7.08} & \best{27.9} & \best{24.1}& \best{16.3}& \best{7.26}& \best{20.6} & \best{15.7}\\
            \arrayrulecolor{black}\hline
            \end{tabular}
            }
        }
     \vspace{-4mm}
    \end{table*}

\begin{table*}[t]
    \caption{Ablation studies~(\S\ref{sec:ABS}) with \textit{audio-goal nav.}~(AGN) and \textit{vision-language nav.}~(VLN) tasks.}
    \label{tab:ablations}
    \hspace{-0.9em}
    \subfloat[\small{multisensory integration} \label{tab:ablation:MTI}]{
        \resizebox{0.317\textwidth}{!}{
        \tablestyle{1pt}{1.1}
        \begin{tabular}[t]{|c|c|c|}
        \hline
        ~ &  { AGN (\textbf{SR}$^\uparrow$)} &  {VLN (\textbf{SR}$^\uparrow$)}\\
        \cline{2-3}
        \multirow{-2}{*}{\tabincell{c}{Modality\\(\S\ref{sec:TP})}} &  {\texttt{seen}/\texttt{unseen}} &  {\texttt{seen}/\texttt{unseen}} \\
        \hline
        {RGB \textit{only}}& 2.5 / 2.1 & 21.1 / 11.2  \\
        \arrayrulecolor{gray}\cdashline{1-3}[1pt/1pt]
        {audio \textit{only}}& 23.1 / 15.9 & 0.3 / 0.2 \\
        \arrayrulecolor{gray}\cdashline{1-3}[1pt/1pt]
        {RGBD \textit{only}}& 2.4 / 2.3 & 21.9 / 12.5 \\
        \arrayrulecolor{gray}\cdashline{1-3}[1pt/1pt]
        RGBD$_{\!_{\!}}$ + $_{\!_{\!}}$audio& 25.3 / 18.7 & 26.5 / 16.3 \\
        \arrayrulecolor{black}\hline
        \end{tabular}
        }
        }
    \hspace{-0.6em}	
    \subfloat[\small{$_{\!}$augmented$_{\!}$ target$_{\!}$ description$_{\!}$ embedding$_{\!}$}\label{tab:ablation:atde}]{%
        \tablestyle{1pt}{1.11}
        \resizebox{0.42\textwidth}{!}{
        \begin{tabular}[t]{|c|c|c|}
        \hline
        ~ &  {AGN (\textbf{SR}$^\uparrow$)} &  {VLN (\textbf{SR}$^\uparrow$)}\\
        \cline{2-3}
        \multirow{-2}{*}{$\bm{G}$ (Eq.~\ref{equ:g})} &  {\texttt{seen}/\texttt{unseen}} &  {\texttt{seen}/\texttt{unseen}} \\
        \hline
        {episodic target only}& 23.1 / 17.2 & 22.7 / 14.1 \\
        \arrayrulecolor{gray}\cdashline{1-3}[1pt/1pt]
        \tabincell{c}{episodic target + \\episodic task embedding} & 24.2 / 17.9 &  25.1 / 15.5 \\
        \arrayrulecolor{gray}\cdashline{1-3}[1pt/1pt]
        {augmented} target des. embed. & 25.3 / 18.7 & 26.5 / 16.3 \\
        \arrayrulecolor{black}\hline
        \end{tabular}
        }}
    \hspace{-0.4em}	
    \subfloat[\small{diversified target parsing}\label{tab:ablation:DTP}]{%
    \resizebox{0.271\textwidth}{!}{
        \tablestyle{1pt}{1.1}
        \begin{tabular}[t]{|c|c|c|}
\hline
~ &  {AGN (\textbf{SR}$^\uparrow$)} &  {VLN (\textbf{SR}$^\uparrow$)}\\
\cline{2-3}
\multirow{-2}{*}{\tabincell{c}{$\bm{Q}_t$\\(Eq.~\ref{equ:q})}} &  {\texttt{seen}/\texttt{unseen}} &  {\texttt{seen}/\texttt{unseen}} \\
\hline
$N_{\!}=_{\!}1$& 21.0 / 15.6& 22.2 / 13.2\\
\arrayrulecolor{gray}\cdashline{1-3}[1pt/1pt]
$N_{\!}=_{\!}3$& 23.8 / 17.8& 25.7 / 15.5\\
\arrayrulecolor{gray}\cdashline{1-3}[1pt/1pt]
$N_{\!}=_{\!}5$& 25.3 / 18.7& 26.5 / 16.3 \\
\arrayrulecolor{gray}\cdashline{1-3}[1pt/1pt]
$N_{\!}=_{\!}7$& 24.9 / 18.3 & 26.2 / 16.2 \\
\arrayrulecolor{black}\hline
\end{tabular}
}}\vfill
\vspace{-.11in}
    \hspace{-0.8em}
    \subfloat[\small{multitask planner}\label{tab:ablation:MTP}]{
        \tablestyle{1pt}{1.1}
        \resizebox{0.278\textwidth}{!}{
        \begin{tabular}[t]{|c|c|c|}
        \hline
        ~ &  {AGN (\textbf{SR}$^\uparrow$)} &  {VLN (\textbf{SR}$^\uparrow$)}\\
        \cline{2-3}
        \multirow{-2}{*}{\tabincell{c}{$F_{\text{MTP}}$\\(Eq.~\ref{eq:MTP})	}} &  {\texttt{seen}/\texttt{unseen}} &  {\texttt{seen}/\texttt{unseen}} \\
        \hline
        separate& 23.2 / 17.4& 25.0 / 15.1 \\
        \arrayrulecolor{gray}\cdashline{1-3}[1pt/1pt]
        $1$-shared& 24.4 / 18.1& 25.8 / 15.7 \\
        \arrayrulecolor{gray}\cdashline{1-3}[1pt/1pt]
        $2$-shared& 25.3 / 18.7& 26.5 / 16.3 \\
        \arrayrulecolor{gray}\cdashline{1-3}[1pt/1pt]
        $3$-shared& 24.8 / 17.9 & 25.9 / 15.5 \\
        \arrayrulecolor{gray}\cdashline{1-3}[1pt/1pt]
        \textit{all}-shared& 24.1 / 17.6& 25.6 / 15.3 \\
        \arrayrulecolor{black}\hline
        \end{tabular}
        }}
        \hspace{-0.35em}
    \subfloat[{reward function}\label{tab:ablation:RF}]{%
        \resizebox{0.374\textwidth}{!}{
        \tablestyle{1pt}{1.09}
        \begin{tabular}[t]{|c|c|c|}
        \hline
         & {AGN (\textbf{SR}$^\uparrow$)} &  {VLN (\textbf{SR}$^\uparrow$)}\\
        \cline{2-3}
        \multirow{-2}{*}{$R$ (\S\ref{sec:MDPPO})} &  {\texttt{seen}/\texttt{unseen}}&  {\texttt{seen}/\texttt{unseen}} \\
        \hline
        $r_{\text{success}\!}$ & 2.1 / 1.5 & 5.5 / 2.3 \\
        \arrayrulecolor{gray}\cdashline{1-3}[1pt/1pt]
        $r_{\text{success}\!_{\!}}$ + $_{_{\!}\!}r_{\text{progress}}$ & 22.5 / 16.3 & 23.8 / 14.7  \\
        \arrayrulecolor{gray}\cdashline{1-3}[1pt/1pt]
        $r_{\text{success}_{\!}\!}$ + $_{_{\!}\!}r_{\text{progress}\!}$ + $_{_{\!}\!}r_{\text{slack}_{\!}}$ & 23.1 / 16.9 & 24.3 / 15.1 \\
        \arrayrulecolor{gray}\cdashline{1-3}[1pt/1pt]
        $r_{\text{success}\!_{\!}}$ + $_{_{\!}\!}r_{\text{slack}\!_{\!}}$ +&&\\
        \specialrule{0em}{-0.5pt}{-1.5pt}
        $_{_{\!}\!}r_{\text{progress}\!_{\!}}$ +$_{_{\!}}r_{\text{explore}}$ &  \multirow{-2}{*}{25.3 / 18.7}  &  \multirow{-2}{*}{26.5 / 16.3} \\
        \arrayrulecolor{black}\hline
        \end{tabular}
        }
    }
        \hspace{-0.55em}
    \subfloat[{multitask learning}\label{tab:ablation:MTL}]{%
           \resizebox{0.3515\textwidth}{!}{
        \tablestyle{1pt}{1.1}
        \begin{tabular}[t]{|c|c|c|}
\hline
~ &  {AGN (\textbf{SR}$^\uparrow$)} &  {VLN (\textbf{SR}$^\uparrow$)}\\
\cline{2-3}
\multirow{-2}{*}{\tabincell{c}{Task}} &  {\texttt{seen}/\texttt{unseen}} &  {\texttt{seen}/\texttt{unseen}} \\
\hline
single task& 22.1 / 15.4 & 23.8 / 14.1 \\
\arrayrulecolor{gray}\cdashline{1-3}[1pt/1pt]
AGN$_{\!}$ + $_{\!}$VLN & 22.7 / 16.1 & 24.4 / 14.9 \\
\arrayrulecolor{gray}\cdashline{1-3}[1pt/1pt]
AGN$_{\!}$ + $_{\!}$VLN$_{\!}$ + $_{\!}$IGN& 24.1 / 17.3 & 25.1 / 15.5 \\
\arrayrulecolor{gray}\cdashline{1-3}[1pt/1pt]
AGN$_{\!}$ + $_{\!}$VLN$_{\!}$ + &&\\
\specialrule{0em}{-0.5pt}{-0.5pt}
IGN$_{\!}$ + $_{\!}$OGN& \multirow{-2}{*}{25.3 / 18.7} &  \multirow{-2}{*}{26.5 / 16.3} \\
\arrayrulecolor{black}\hline
\end{tabular}
}
}
\vspace{-5mm}
\end{table*}

\subsection{Ablative Study}\label{sec:ABS}
\vspace{-3pt}
To thoroughly test the efficacy of crucial components of \textsc{Vienna}, we conduct a series of~diagnostic studies on \textit{vision-language nav.} and \textit{audio-goal nav.} tasks. The results are summarized in Table~\ref{tab:ablations}.

\noindent\textbf{Multisensory Integration.} Agents in \textbf{\texttt{VXN}} are equipped with a multimodal sensor so as to find the target by both seeing and hearing and make our navigation setting closer the real-world. We first study the influence of different sensory signals (\ie, RGB, depth, audio) by training$_{\!}$ \textsc{Vienna}$_{\!}$ with$_{\!}$~varying sensory$_{\!}$ modalities.$_{\!}$ As$_{\!}$ shown$_{\!}$ in$_{\!}$ Table$_{\!}$~\ref{tab:ablation:MTI},$_{\!}$ fusing$_{\!}$ multimodal$_{\!}$ sensory$_{\!}$ cues$_{\!}$ (\ie,$_{\!}$ RGBD$_{\!}$ +$_{\!}$ audio)$_{\!}$ is$_{\!}$ more favored. For example, in VLN, although considering$_{\!}$ audio$_{\!}$ alone$_{\!}$ brings$_{\!}$ poor$_{\!}$ performance$_{\!}$, supplementing RGBD perception with audio yields notable improvements. This suggests  audio is complementary to visual sensory in capturing physical and semantic properties of environments.

\noindent\textbf{Augmented Target Description Embedding.} To better master cross-task knowledge, \textsc{Vienna}$_{\!}$ augments its episodic targets with all the four learnable task embeddings $\bm{\tau}_{{I, A, T, L}\!}$  (\textit{cf.}$_{\!}$~Eq.~\ref{equ:g}). {We compare this design against two variants in Table$_{\!}$~\ref{tab:ablation:atde}, and find such a strategy is conducive to the performance.} This is because, through end-to-end training, $\bm{\tau}_{{I, A, T, L}\!}$ carry task-specific knowledge, \eg, $\bm{\tau}_{A}$ is associated with some discriminative audio clips; $\bm{\tau}_{I}$ focuses on essential visual landmarks. By taking $\bm{\tau}_{{I, A, T, L}\!}$ together, \textsc{Vienna}$_{\!}$ use key knowledge of different tasks in single task episodes.

\noindent\textbf{Diversified Target Parsing.} We online parse the augmented target description embedding $\bm{G}$ into $N$
target embeddings $\bm{Q}_t\!=\![\bm{q}^1_t,\cdots,\bm{q}^N_t]$ (\textit{cf.}$_{\!}$~Eq.$_{\!}$~\ref{equ:q}), to achieve vivid and diversified interpretations of $\bm{G}$. In Table~\ref{tab:ablation:DTP}, we present evaluation scores with different numbers of generated target embeddings, \ie, $N=1,3,5,7$. As can be seen, diversified target parsing indeed boots navigation performance.

\noindent\textbf{Multitask Planner.} Several variants of multitask planner $f_{\textsc{Mtp}}$ (\textit{cf}.$_{\!}$~Eq.$_{\!}$~\ref{eq:MTP}) are compared in Table$_{\!}$~\ref{tab:ablation:MTP}. {The two-layer shared trunk design is adopted, due to its relatively better performance.}

\noindent\textbf{Reward Function.} Next we examine the design of our reward function (\S\ref{sec:MDPPO}). As seen in Table$_{\!}$~\ref{tab:ablation:RF}, each reward term is indeed useful and combining all the four terms leads to the best performance.

\noindent\textbf{Multitask Learning.} Table$_{\!}$~\ref{tab:ablation:MTL} reveals the value of training \textsc{Vienna} on multiple tasks: training with more navigation tasks improves both performance and generalizability.
Compared to a composition of four single-task models, multi-task \textsc{Vienna} also greatly reduces the model size: 101M $\!\rightarrow\!$ 31M.

\vspace{-5pt}
\section{Conclusion}
\vspace{-3pt}
In this work, we present \textbf{\texttt{VXN}}, a large-scale 3D indoor dataset~for multimodal, multitask navigation in continuous and audiovisual complex environments. Further, we devise$_{\!}$ \textsc{Vienna}, a powerful agent that simultaneously learns four famous navigation tasks within a single unified parsing-and-query scheme. We empirically show that, through a fully attentive architecture, \textsc{Vienna} is able to mine and utilize cross-task knowledge to enhance the performance on all the tasks. These efforts move us closer to a community goal of general-purpose robots capable of fulfilling a multitude of tasks.

\medskip
{\small
\bibliographystyle{unsrt}
\bibliography{egbib}

\begin{thebibliography}{100}

\bibitem{kim1999symbolic}
Dongsung Kim and Ramakant Nevatia.
\newblock Symbolic navigation with a generic map.
\newblock {\em Autonomous Robots}, 6(1):69--88, 1999.

\bibitem{chaplot2020object}
Devendra~Singh Chaplot, Dhiraj Gandhi, Abhinav Gupta, and Ruslan Salakhutdinov.
\newblock Object goal navigation using goal-oriented semantic exploration.
\newblock In {\em NeurIPS}, 2020.

\bibitem{chaplot2020neural}
Devendra~Singh Chaplot, Ruslan Salakhutdinov, Abhinav Gupta, and Saurabh Gupta.
\newblock Neural topological {SLAM} for visual navigation.
\newblock In {\em CVPR}, 2020.

\bibitem{striegnitz2011report}
Kristina Striegnitz, Alexandre Denis, Andrew Gargett, Konstantina Garoufi,
  Alexander Koller, and Mari{\"e}t Theune.
\newblock Report on the second challenge on generating instructions in virtual
  environments (give-2.5).
\newblock In {\em European Workshop on Natural Language Generation}, 2011.

\bibitem{zhu2017target}
Yuke Zhu, Roozbeh Mottaghi, Eric Kolve, Joseph~J Lim, Abhinav Gupta,
  Li~Fei-Fei, and Ali Farhadi.
\newblock Target-driven visual navigation in indoor scenes using deep
  reinforcement learning.
\newblock In {\em ICRA}, 2017.

\bibitem{chen2020soundspaces}
Changan Chen, Unnat Jain, Carl Schissler, Sebastia Vicenc~Amengual Gari, Ziad
  Al-Halah, Vamsi~Krishna Ithapu, Philip Robinson, and Kristen Grauman.
\newblock Soundspaces: Audio-visual navigation in 3d environments.
\newblock In {\em ECCV}, 2020.

\bibitem{yang2018visual}
Wei Yang, Xiaolong Wang, Ali Farhadi, Abhinav Gupta, and Roozbeh Mottaghi.
\newblock Visual semantic navigation using scene priors.
\newblock In {\em ICLR}, 2019.

\bibitem{anderson2018vision}
Peter Anderson, Qi~Wu, Damien Teney, Jake Bruce, Mark Johnson, Niko
  S{\"u}nderhauf, Ian Reid, Stephen Gould, and Anton van~den Hengel.
\newblock Vision-and-language navigation: Interpreting visually-grounded
  navigation instructions in real environments.
\newblock In {\em CVPR}, 2018.

\bibitem{zhang2018overview}
Yu~Zhang and Qiang Yang.
\newblock An overview of multi-task learning.
\newblock {\em National Science Review}, 5(1):30--43, 2018.

\bibitem{meredith1986visual}
M~Alex Meredith and Barry~E Stein.
\newblock Visual, auditory, and somatosensory convergence on cells in superior
  colliculus results in multisensory integration.
\newblock {\em Journal of neurophysiology}, 56(3):640--662, 1986.

\bibitem{kording2007causal}
Konrad~P K{\"o}rding, Ulrik Beierholm, Wei~Ji Ma, Steven Quartz, Joshua~B
  Tenenbaum, and Ladan Shams.
\newblock Causal inference in multisensory perception.
\newblock {\em PLoS one}, 2(9):e943, 2007.

\bibitem{caruana1997multitask}
Rich Caruana.
\newblock Multitask learning.
\newblock {\em Machine learning}, 28(1):41--75, 1997.

\bibitem{crawshaw2020multi}
Michael Crawshaw.
\newblock Multi-task learning with deep neural networks: A survey.
\newblock {\em arXiv preprint arXiv:2009.09796}, 2020.

\bibitem{zhang2021survey}
Yu~Zhang and Qiang Yang.
\newblock A survey on multi-task learning.
\newblock {\em TKDE}, 2021.

\bibitem{2019Habitat}
Manolis Savva, Abhishek Kadian, Oleksandr Maksymets, Yili Zhao, Erik Wijmans,
  Bhavana Jain, Julian Straub, Jia Liu, Vladlen Koltun, Jitendra Malik, et~al.
\newblock Habitat: A platform for embodied ai research.
\newblock In {\em ICCV}, 2019.

\bibitem{vaswani2017attention}
Ashish Vaswani, Noam Shazeer, Niki Parmar, Jakob Uszkoreit, Llion Jones,
  Aidan~N Gomez, Lukasz Kaiser, and Illia Polosukhin.
\newblock Attention is all you need.
\newblock In {\em NeurIPS}, 2017.

\bibitem{giralt1979multi}
Georges Giralt, Ralph Sobek, and Raja Chatila.
\newblock A multi-level planning and navigation system for a mobile robot: a
  first approach to hilare.
\newblock In {\em IJCAI}, 1979.

\bibitem{armeni20163d}
Iro Armeni, Ozan Sener, Amir~R Zamir, Helen Jiang, Ioannis Brilakis, Martin
  Fischer, and Silvio Savarese.
\newblock {3D} semantic parsing of large-scale indoor spaces.
\newblock In {\em CVPR}, 2016.

\bibitem{chang2017matterport3d}
Angel Chang, Angela Dai, Thomas Funkhouser, Maciej Halber, Matthias Niebner,
  Manolis Savva, Shuran Song, Andy Zeng, and Yinda Zhang.
\newblock {Matterport3D}: Learning from rgb-d data in indoor environments.
\newblock In {\em 3DV}, 2018.

\bibitem{wu2018building}
Yi~Wu, Yuxin Wu, Georgia Gkioxari, and Yuandong Tian.
\newblock Building generalizable agents with a realistic and rich 3d
  environment.
\newblock {\em arXiv preprint arXiv:1801.02209}, 2018.

\bibitem{song2017semantic}
Shuran Song, Fisher Yu, Andy Zeng, Angel~X Chang, Manolis Savva, and Thomas
  Funkhouser.
\newblock Semantic scene completion from a single depth image.
\newblock In {\em CVPR}, 2017.

\bibitem{kolve2017ai2}
Eric Kolve, Roozbeh Mottaghi, Winson Han, Eli VanderBilt, Luca Weihs, Alvaro
  Herrasti, Daniel Gordon, Yuke Zhu, Abhinav Gupta, and Ali Farhadi.
\newblock Ai2-thor: An interactive 3d environment for visual ai.
\newblock {\em arXiv preprint arXiv:1712.05474}, 2017.

\bibitem{xia2018gibson}
Fei Xia, Amir~R Zamir, Zhiyang He, Alexander Sax, Jitendra Malik, and Silvio
  Savarese.
\newblock Gibson env: Real-world perception for embodied agents.
\newblock In {\em CVPR}, 2018.

\bibitem{deitke2020robothor}
Matt Deitke, Winson Han, Alvaro Herrasti, Aniruddha Kembhavi, Eric Kolve,
  Roozbeh Mottaghi, Jordi Salvador, Dustin Schwenk, Eli VanderBilt, Matthew
  Wallingford, et~al.
\newblock Robothor: An open simulation-to-real embodied ai platform.
\newblock In {\em CVPR}, 2020.

\bibitem{krantz2020navgraph}
Jacob Krantz, Erik Wijmans, Arjun Majumdar, Dhruv Batra, and Stefan Lee.
\newblock Beyond the nav-graph: Vision-and-language navigation in continuous
  environments.
\newblock In {\em ECCV}, 2020.

\bibitem{das2018embodied}
Abhishek Das, Samyak Datta, Georgia Gkioxari, Stefan Lee, Devi Parikh, and
  Dhruv Batra.
\newblock Embodied question answering.
\newblock In {\em CVPR}, 2018.

\bibitem{2019Vision}
Jesse Thomason, Michael Murray, Maya Cakmak, and Luke Zettlemoyer.
\newblock Vision-and-dialog navigation.
\newblock In {\em CoRL}, 2019.

\bibitem{2018Vision}
Khanh Nguyen, Debadeepta Dey, Chris Brockett, and Bill Dolan.
\newblock Vision-based navigation with language-based assistance via imitation
  learning with indirect intervention.
\newblock In {\em CVPR}, 2018.

\bibitem{2019Help}
Khanh Nguyen and Hal Daum{\'e}~III.
\newblock Help, anna! visual navigation with natural multimodal assistance via
  retrospective curiosity-encouraging imitation learning.
\newblock In {\em EMNLP-IJCNLP}, 2019.

\bibitem{wang2021collaborative}
Haiyang Wang, Wenguan Wang, Xizhou Zhu, Jifeng Dai, and Liwei Wang.
\newblock Collaborative visual navigation.
\newblock {\em arXiv preprint arXiv:2107.01151}, 2021.

\bibitem{mirowski2017learning}
Piotr Mirowski, Razvan Pascanu, Fabio Viola, Hubert Soyer, Andrew~J Ballard,
  Andrea Banino, Misha Denil, Ross Goroshin, Laurent Sifre, Koray Kavukcuoglu,
  et~al.
\newblock Learning to navigate in complex environments.
\newblock In {\em ICLR}, 2017.

\bibitem{gupta2017cognitive}
Saurabh Gupta, James Davidson, Sergey Levine, Rahul Sukthankar, and Jitendra
  Malik.
\newblock Cognitive mapping and planning for visual navigation.
\newblock In {\em CVPR}, 2017.

\bibitem{parisotto2017neural}
Emilio Parisotto and Ruslan Salakhutdinov.
\newblock Neural map: Structured memory for deep reinforcement learning.
\newblock In {\em ICLR}, 2019.

\bibitem{zhang2017neural}
Jingwei Zhang, Lei Tai, Joschka Boedecker, Wolfram Burgard, and Ming Liu.
\newblock Neural slam: Learning to explore with external memory.
\newblock {\em arXiv preprint arXiv:1706.09520}, 2017.

\bibitem{wu2019bayesian}
Yi~Wu, Yuxin Wu, Aviv Tamar, Stuart Russell, Georgia Gkioxari, and Yuandong
  Tian.
\newblock Bayesian relational memory for semantic visual navigation.
\newblock In {\em ICCV}, 2019.

\bibitem{chaplot2018active}
Devendra~Singh Chaplot, Emilio Parisotto, and Ruslan Salakhutdinov.
\newblock Active neural localization.
\newblock In {\em ICLR}, 2018.

\bibitem{savinov2018semi}
Nikolay Savinov, Alexey Dosovitskiy, and Vladlen Koltun.
\newblock Semi-parametric topological memory for navigation.
\newblock In {\em ICLR}, 2018.

\bibitem{chaplot2020learning}
Devendra~Singh Chaplot, Dhiraj Gandhi, Saurabh Gupta, Abhinav Gupta, and Ruslan
  Salakhutdinov.
\newblock Learning to explore using active neural slam.
\newblock In {\em ICLR}, 2020.

\bibitem{wang2021structured}
Hanqing Wang, Wenguan Wang, Wei Liang, Caiming Xiong, and Jianbing Shen.
\newblock Structured scene memory for vision-language navigation.
\newblock In {\em CVPR}, 2021.

\bibitem{Chen_2021_CVPR}
Kevin Chen, Junshen~K. Chen, Jo~Chuang, Marynel Vazquez, and Silvio Savarese.
\newblock Topological planning with transformers for vision-and-language
  navigation.
\newblock In {\em CVPR}, 2021.

\bibitem{zhao2022target}
Yusheng Zhao, Jinyu Chen, Chen Gao, Wenguan Wang, Lirong Yang, Haibing Ren,
  Huaxia Xia, and Si~Liu.
\newblock Target-driven structured transformer planner for vision-language
  navigation.
\newblock In {\em ACMMM}, 2022.

\bibitem{lee2018gated}
Lisa Lee, Emilio Parisotto, Devendra~Singh Chaplot, Eric Xing, and Ruslan
  Salakhutdinov.
\newblock Gated path planning networks.
\newblock In {\em ICML}, 2018.

\bibitem{deng2020evolving}
Zhiwei Deng, Karthik Narasimhan, and Olga Russakovsky.
\newblock Evolving graphical planner: Contextual global planning for
  vision-and-language navigation.
\newblock {\em arXiv preprint arXiv:2007.05655}, 2020.

\bibitem{wang2020active}
Hanqing Wang, Wenguan Wang, Tianmin Shu, Wei Liang, and Jianbing Shen.
\newblock Active visual information gathering for vision-language navigation.
\newblock In {\em ECCV}, 2020.

\bibitem{hu2019you}
Ronghang Hu, Daniel Fried, Anna Rohrbach, Dan Klein, Trevor Darrell, and Kate
  Saenko.
\newblock Are you looking? grounding to multiple modalities in
  vision-and-language navigation.
\newblock In {\em ACL}, 2019.

\bibitem{qi2020object}
Yuankai Qi, Zizheng Pan, Shengping Zhang, Anton van~den Hengel, and Qi~Wu.
\newblock Object-and-action aware model for visual language navigation.
\newblock In {\em ECCV}, 2020.

\bibitem{wang2020environment}
Xin~Eric Wang, Vihan Jain, Eugene Ie, William~Yang Wang, Zornitsa Kozareva, and
  Sujith Ravi.
\newblock Environment-agnostic multitask learning for natural language grounded
  navigation.
\newblock In {\em ECCV}, 2020.

\bibitem{Hong_2021_CVPR}
Yicong Hong, Qi~Wu, Yuankai Qi, Cristian Rodriguez-Opazo, and Stephen Gould.
\newblock Vln bert: A recurrent vision-and-language bert for navigation.
\newblock In {\em CVPR}, 2021.

\bibitem{fried2018speaker}
Daniel Fried, Ronghang Hu, Volkan Cirik, Anna Rohrbach, Jacob Andreas,
  Louis-Philippe Morency, Taylor Berg-Kirkpatrick, Kate Saenko, Dan Klein, and
  Trevor Darrell.
\newblock Speaker-follower models for vision-and-language navigation.
\newblock In {\em NeurIPS}, 2018.

\bibitem{tan2019learning}
Hao Tan, Licheng Yu, and Mohit Bansal.
\newblock Learning to navigate unseen environments: Back translation with
  environmental dropout.
\newblock In {\em NAACL}, 2019.

\bibitem{fu2019counterfactual}
Tsu-Jui Fu, Xin Wang, Matthew Peterson, Scott Grafton, Miguel Eckstein, and
  William~Yang Wang.
\newblock Counterfactual vision-and-language navigation via adversarial path
  sampling.
\newblock In {\em ECCV}, 2020.

\bibitem{majumdar2020improving}
Arjun Majumdar, Ayush Shrivastava, Stefan Lee, Peter Anderson, Devi Parikh, and
  Dhruv Batra.
\newblock Improving vision-and-language navigation with image-text pairs from
  the web.
\newblock In {\em ECCV}, 2020.

\bibitem{hao2020prevalent}
Weituo Hao, Chunyuan Li, Xiujun Li, Lawrence Carin, and Jianfeng Gao.
\newblock Towards learning a generic agent for vision-and-language navigation
  via pre-training.
\newblock In {\em CVPR}, 2020.

\bibitem{wang2022counterfactual}
Hanqing Wang, Wei Liang, Jianbing Shen, Luc Van~Gool, and Wenguan Wang.
\newblock Counterfactual cycle-consistent learning for instruction following
  and generation in vision-language navigation.
\newblock In {\em CVPR}, 2022.

\bibitem{gao2021room}
Chen Gao, Jinyu Chen, Si~Liu, Luting Wang, Qiong Zhang, and Qi~Wu.
\newblock Room-and-object aware knowledge reasoning for remote embodied
  referring expression.
\newblock In {\em CVPR}, 2021.

\bibitem{ruder2017overview}
Sebastian Ruder.
\newblock An overview of multi-task learning in deep neural networks.
\newblock {\em arXiv preprint arXiv:1706.05098}, 2017.

\bibitem{wang2018attentive}
Wenguan Wang, Yuanlu Xu, Jianbing Shen, and Song-Chun Zhu.
\newblock Attentive fashion grammar network for fashion landmark detection and
  clothing category classification.
\newblock In {\em CVPR}, 2018.

\bibitem{wang2019salient}
Wenguan Wang, Shuyang Zhao, Jianbing Shen, Steven~CH Hoi, and Ali Borji.
\newblock Salient object detection with pyramid attention and salient edges.
\newblock In {\em CVPR}, 2019.

\bibitem{lu202012}
Jiasen Lu, Vedanuj Goswami, Marcus Rohrbach, Devi Parikh, and Stefan Lee.
\newblock 12-in-1: Multi-task vision and language representation learning.
\newblock In {\em CVPR}, 2020.

\bibitem{collobert2008unified}
Ronan Collobert and Jason Weston.
\newblock A unified architecture for natural language processing: Deep neural
  networks with multitask learning.
\newblock In {\em ICML}, 2008.

\bibitem{zhang2014facial}
Zhanpeng Zhang, Ping Luo, Chen~Change Loy, and Xiaoou Tang.
\newblock Facial landmark detection by deep multi-task learning.
\newblock In {\em ECCV}, 2014.

\bibitem{misra2016cross}
Ishan Misra, Abhinav Shrivastava, Abhinav Gupta, and Martial Hebert.
\newblock Cross-stitch networks for multi-task learning.
\newblock In {\em CVPR}, 2016.

\bibitem{xu2018pad}
Dan Xu, Wanli Ouyang, Xiaogang Wang, and Nicu Sebe.
\newblock Pad-net: Multi-tasks guided prediction-and-distillation network for
  simultaneous depth estimation and scene parsing.
\newblock In {\em CVPR}, 2018.

\bibitem{strezoski2019many}
Gjorgji Strezoski, Nanne~van Noord, and Marcel Worring.
\newblock Many task learning with task routing.
\newblock In {\em ICCV}, 2019.

\bibitem{kendall2018multi}
Alex Kendall, Yarin Gal, and Roberto Cipolla.
\newblock Multi-task learning using uncertainty to weigh losses for scene
  geometry and semantics.
\newblock In {\em CVPR}, 2018.

\bibitem{chen2018gradnorm}
Zhao Chen, Vijay Badrinarayanan, Chen-Yu Lee, and Andrew Rabinovich.
\newblock Gradnorm: Gradient normalization for adaptive loss balancing in deep
  multitask networks.
\newblock In {\em ICML}, 2018.

\bibitem{guo2018dynamic}
Michelle Guo, Albert Haque, De-An Huang, Serena Yeung, and Li~Fei-Fei.
\newblock Dynamic task prioritization for multitask learning.
\newblock In {\em ECCV}, 2018.

\bibitem{duong2015low}
Long Duong, Trevor Cohn, Steven Bird, and Paul Cook.
\newblock Low resource dependency parsing: Cross-lingual parameter sharing in a
  neural network parser.
\newblock In {\em IJCNLP}, 2015.

\bibitem{sanh2019hierarchical}
Victor Sanh, Thomas Wolf, and Sebastian Ruder.
\newblock A hierarchical multi-task approach for learning embeddings from
  semantic tasks.
\newblock In {\em AAAI}, 2019.

\bibitem{sener2018multi}
Ozan Sener and Vladlen Koltun.
\newblock Multi-task learning as multi-objective optimization.
\newblock In {\em NeurIPS}, 2018.

\bibitem{bingel2017identifying}
Joachim Bingel and Anders S{\o}gaard.
\newblock Identifying beneficial task relations for multi-task learning in deep
  neural networks.
\newblock In {\em ACL}, 2017.

\bibitem{zamir2018taskonomy}
Amir~R Zamir, Alexander Sax, William Shen, Leonidas~J Guibas, Jitendra Malik,
  and Silvio Savarese.
\newblock Taskonomy: Disentangling task transfer learning.
\newblock In {\em CVPR}, 2018.

\bibitem{dwivedi2019representation}
Kshitij Dwivedi and Gemma Roig.
\newblock Representation similarity analysis for efficient task taxonomy \&
  transfer learning.
\newblock In {\em CVPR}, 2019.

\bibitem{yang2017multi}
Zhaoyang Yang, Kathryn~E Merrick, Hussein~A Abbass, and Lianwen Jin.
\newblock Multi-task deep reinforcement learning for continuous action control.
\newblock In {\em IJCAI}, 2017.

\bibitem{teh2017distral}
Yee~Whye Teh, Victor Bapst, Wojciech~Marian Czarnecki, John Quan, James
  Kirkpatrick, Raia Hadsell, Nicolas Heess, and Razvan Pascanu.
\newblock Distral: robust multitask reinforcement learning.
\newblock In {\em NeurIPS}, 2017.

\bibitem{espeholt2018impala}
Lasse Espeholt, Hubert Soyer, Remi Munos, Karen Simonyan, Vlad Mnih, Tom Ward,
  Yotam Doron, Vlad Firoiu, Tim Harley, Iain Dunning, et~al.
\newblock Impala: Scalable distributed deep-rl with importance weighted
  actor-learner architectures.
\newblock In {\em ICML}, 2018.

\bibitem{pinto2017learning}
Lerrel Pinto and Abhinav Gupta.
\newblock Learning to push by grasping: Using multiple tasks for effective
  learning.
\newblock In {\em ICRA}, 2017.

\bibitem{zeng2018learning}
Andy Zeng, Shuran Song, Stefan Welker, Johnny Lee, Alberto Rodriguez, and
  Thomas Funkhouser.
\newblock Learning synergies between pushing and grasping with self-supervised
  deep reinforcement learning.
\newblock In {\em IROS}, 2018.

\bibitem{hessel2019multi}
Matteo Hessel, Hubert Soyer, Lasse Espeholt, Wojciech Czarnecki, Simon Schmitt,
  and Hado van Hasselt.
\newblock Multi-task deep reinforcement learning with popart.
\newblock In {\em AAAI}, 2019.

\bibitem{d2020sharing}
Carlo D'Eramo, Davide Tateo, Andrea Bonarini, Marcello Restelli, and Jan
  Peters.
\newblock Sharing knowledge in multi-task deep reinforcement learning.
\newblock In {\em ICLR}, 2020.

\bibitem{xu2020knowledge}
Zhiyuan Xu, Kun Wu, Zhengping Che, Jian Tang, and Jieping Ye.
\newblock Knowledge transfer in multi-task deep reinforcement learning for
  continuous control.
\newblock {\em arXiv preprint arXiv:2010.07494}, 2020.

\bibitem{heess2016learning}
Nicolas Heess, Greg Wayne, Yuval Tassa, Timothy Lillicrap, Martin Riedmiller,
  and David Silver.
\newblock Learning and transfer of modulated locomotor controllers.
\newblock {\em arXiv preprint arXiv:1610.05182}, 2016.

\bibitem{devin2017learning}
Coline Devin, Abhishek Gupta, Trevor Darrell, Pieter Abbeel, and Sergey Levine.
\newblock Learning modular neural network policies for multi-task and
  multi-robot transfer.
\newblock In {\em ICRA}, 2017.

\bibitem{rusu2015policy}
Andrei~A Rusu, Sergio~Gomez Colmenarejo, Caglar Gulcehre, Guillaume Desjardins,
  James Kirkpatrick, Razvan Pascanu, Volodymyr Mnih, Koray Kavukcuoglu, and
  Raia Hadsell.
\newblock Policy distillation.
\newblock In {\em ICLR}, 2015.

\bibitem{parisotto2016actor}
Emilio Parisotto, Lei~Jimmy Ba, and Ruslan Salakhutdinov.
\newblock Actor-mimic: Deep multitask and transfer reinforcement learning.
\newblock In {\em ICLR}, 2016.

\bibitem{beattie2016deepmind}
Charles Beattie, Joel~Z Leibo, Denis Teplyashin, Tom Ward, Marcus Wainwright,
  Heinrich K{\"u}ttler, Andrew Lefrancq, Simon Green, V{\'\i}ctor Vald{\'e}s,
  Amir Sadik, et~al.
\newblock Deepmind lab.
\newblock {\em arXiv preprint arXiv:1612.03801}, 2016.

\bibitem{bellemare2013arcade}
Marc~G Bellemare, Yavar Naddaf, Joel Veness, and Michael Bowling.
\newblock The arcade learning environment: An evaluation platform for general
  agents.
\newblock {\em Journal of Artificial Intelligence Research}, 47:253--279, 2013.

\bibitem{yu2020meta}
Tianhe Yu, Deirdre Quillen, Zhanpeng He, Ryan Julian, Karol Hausman, Chelsea
  Finn, and Sergey Levine.
\newblock Meta-world: A benchmark and evaluation for multi-task and meta
  reinforcement learning.
\newblock In {\em Conference on Robot Learning}, 2020.

\bibitem{chaplot2019embodied}
Devendra~Singh Chaplot, Lisa Lee, Ruslan Salakhutdinov, Devi Parikh, and Dhruv
  Batra.
\newblock Embodied multimodal multitask learning.
\newblock {\em arXiv preprint arXiv:1902.01385}, 2019.

\bibitem{mirowski2016learning}
Piotr Mirowski, Razvan Pascanu, Fabio Viola, Hubert Soyer, Andrew~J Ballard,
  Andrea Banino, Misha Denil, Ross Goroshin, Laurent Sifre, Koray Kavukcuoglu,
  et~al.
\newblock Learning to navigate in complex environments.
\newblock In {\em ICLR}, 2017.

\bibitem{gordon2019splitnet}
Daniel Gordon, Abhishek Kadian, Devi Parikh, Judy Hoffman, and Dhruv Batra.
\newblock Splitnet: Sim2sim and task2task transfer for embodied visual
  navigation.
\newblock In {\em ICCV}, 2019.

\bibitem{pathak2017curiosity}
Deepak Pathak, Pulkit Agrawal, Alexei~A Efros, and Trevor Darrell.
\newblock Curiosity-driven exploration by self-supervised prediction.
\newblock In {\em ICML}, 2017.

\bibitem{gregor2019shaping}
Karol Gregor, Danilo Jimenez~Rezende, Frederic Besse, Yan Wu, Hamza Merzic, and
  Aaron van~den Oord.
\newblock Shaping belief states with generative environment models for rl.
\newblock In {\em NeurIPS}, 2019.

\bibitem{ye2020auxiliary}
Joel Ye, Dhruv Batra, Erik Wijmans, and Abhishek Das.
\newblock Auxiliary tasks speed up learning pointgoal navigation.
\newblock {\em arXiv preprint arXiv:2007.04561}, 2020.

\bibitem{ma2019self}
Chih-Yao Ma, Jiasen Lu, Zuxuan Wu, Ghassan AlRegib, Zsolt Kira, Richard Socher,
  and Caiming Xiong.
\newblock Self-monitoring navigation agent via auxiliary progress estimation.
\newblock In {\em ICLR}, 2019.

\bibitem{zhu2019vision}
Fengda Zhu, Yi~Zhu, Xiaojun Chang, and Xiaodan Liang.
\newblock Vision-language navigation with self-supervised auxiliary reasoning
  tasks.
\newblock In {\em CVPR}, 2020.

\bibitem{fang2019scene}
Kuan Fang, Alexander Toshev, Li~Fei-Fei, and Silvio Savarese.
\newblock Scene memory transformer for embodied agents in long-horizon tasks.
\newblock In {\em CVPR}, 2019.

\bibitem{landi2020perceive}
Federico Landi, Lorenzo Baraldi, Marcella Cornia, Massimiliano Corsini, and
  Rita Cucchiara.
\newblock Perceive, transform, and act: Multi-modal attention networks for
  vision-and-language navigation.
\newblock {\em arXiv preprint arXiv:1911.12377}, 2020.

\bibitem{chen2021semantic}
Changan Chen, Ziad Al-Halah, and Kristen Grauman.
\newblock Semantic audio-visual navigation.
\newblock In {\em CVPR}, 2021.

\bibitem{du2021vtnet}
Heming Du, Xin Yu, and Liang Zheng.
\newblock Vtnet: Visual transformer network for object goal navigation.
\newblock {\em arXiv preprint arXiv:2105.09447}, 2021.

\bibitem{pashevich2021episodic}
Alexander Pashevich, Cordelia Schmid, and Chen Sun.
\newblock Episodic transformer for vision-and-language navigation.
\newblock {\em arXiv preprint arXiv:2105.06453}, 2021.

\bibitem{kaiser2017one}
Lukasz Kaiser, Aidan~N Gomez, Noam Shazeer, Ashish Vaswani, Niki Parmar, Llion
  Jones, and Jakob Uszkoreit.
\newblock One model to learn them all.
\newblock {\em arXiv preprint arXiv:1706.05137}, 2017.

\bibitem{lu2019vilbert}
Jiasen Lu, Dhruv Batra, Devi Parikh, and Stefan Lee.
\newblock Vilbert: Pretraining task-agnostic visiolinguistic representations
  for vision-and-language tasks.
\newblock {\em arXiv preprint arXiv:1908.02265}, 2019.

\bibitem{pramanik2019omninet}
Subhojeet Pramanik, Priyanka Agrawal, and Aman Hussain.
\newblock Omninet: A unified architecture for multi-modal multi-task learning.
\newblock {\em arXiv preprint arXiv:1907.07804}, 2019.

\bibitem{hu2021unit}
Ronghang Hu and Amanpreet Singh.
\newblock Unit: Multimodal multitask learning with a unified transformer.
\newblock {\em arXiv preprint arXiv:2102.10772}, 2021.

\bibitem{hao2020towards}
Weituo Hao, Chunyuan Li, Xiujun Li, Lawrence Carin, and Jianfeng Gao.
\newblock Towards learning a generic agent for vision-and-language navigation
  via pre-training.
\newblock In {\em CVPR}, 2020.

\bibitem{Li_2020_CVPR}
Juncheng Li, Xin Wang, Siliang Tang, Haizhou Shi, Fei Wu, Yueting Zhuang, and
  William~Yang Wang.
\newblock Unsupervised reinforcement learning of transferable meta-skills for
  embodied navigation.
\newblock In {\em CVPR}, 2020.

\bibitem{guhur2021airbert}
Pierre-Louis Guhur, Makarand Tapaswi, Shizhe Chen, Ivan Laptev, and Cordelia
  Schmid.
\newblock {Airbert: In-domain Pretraining for Vision-and-Language Navigation}.
\newblock In {\em ICCV}, 2021.

\bibitem{batra2020objectnav}
Dhruv Batra, Aaron Gokaslan, Aniruddha Kembhavi, Oleksandr Maksymets, Roozbeh
  Mottaghi, Manolis Savva, Alexander Toshev, and Erik Wijmans.
\newblock Object{N}av {R}evisited: {O}n {E}valuation of {E}mbodied {A}gents
  {N}avigating to {O}bjects.
\newblock {\em arXiv preprint arXiv:2006.13171}, 2020.

\bibitem{habitat}
Habitat image nav repository.
\newblock \url{https://github.com/facebookresearch/habitat-lab/pull/333}.

\bibitem{2018Binaural}
Z.~Markus, S.~Christian, and H.~Robert.
\newblock Binaural rendering of ambisonic signals by head-related impulse
  response time alignment and a diffuseness constraint.
\newblock {\em The Journal of the Acoustical Society of America},
  143(6):3616--3627, 2018.

\bibitem{kearney2009approximation}
Gavin Kearney, Claire Masterson, Stephen Adams, and Frank Boland.
\newblock Approximation of binaural room impulse responses.
\newblock In {\em ISSC}, 2009.

\bibitem{sakoe1978dynamic}
Hiroaki Sakoe and Seibi Chiba.
\newblock Dynamic programming algorithm optimization for spoken word
  recognition.
\newblock {\em IEEE Transactions on Acoustics, Speech, and Signal processing},
  26(1):43--49, 1978.

\bibitem{anderson2018evaluation}
Peter Anderson, Angel Chang, Devendra~Singh Chaplot, Alexey Dosovitskiy,
  Saurabh Gupta, Vladlen Koltun, Jana Kosecka, Jitendra Malik, Roozbeh
  Mottaghi, Manolis Savva, et~al.
\newblock On evaluation of embodied navigation agents.
\newblock {\em arXiv preprint arXiv:1807.06757}, 2018.

\bibitem{chen2019touchdown}
Howard Chen, Alane Suhr, Dipendra Misra, Noah Snavely, and Yoav Artzi.
\newblock Touchdown: Natural language navigation and spatial reasoning in
  visual street environments.
\newblock In {\em CVPR}, 2019.

\bibitem{bellman1957markovian}
Richard Bellman.
\newblock A markovian decision process.
\newblock {\em Journal of mathematics and mechanics}, 6(5):679--684, 1957.

\bibitem{heess2017emergence}
Nicolas Heess, Dhruva TB, Srinivasan Sriram, Jay Lemmon, Josh Merel, Greg
  Wayne, Yuval Tassa, Tom Erez, Ziyu Wang, SM~Eslami, et~al.
\newblock Emergence of locomotion behaviours in rich environments.
\newblock {\em arXiv preprint arXiv:1707.02286}, 2017.

\bibitem{ye2021auxiliary}
Joel Ye, Dhruv Batra, Abhishek Das, and Erik Wijmans.
\newblock Auxiliary tasks and exploration enable objectnav.
\newblock {\em arXiv preprint arXiv:2104.04112}, 2021.

\bibitem{schulman2017proximal}
John Schulman, Filip Wolski, Prafulla Dhariwal, Alec Radford, and Oleg Klimov.
\newblock Proximal policy optimization algorithms.
\newblock {\em arXiv preprint arXiv:1707.06347}, 2017.

\bibitem{ILSVRC15}
Olga Russakovsky, Jia Deng, Hao Su, Jonathan Krause, Sanjeev Satheesh, Sean Ma,
  Zhiheng Huang, Andrej Karpathy, Aditya Khosla, Michael Bernstein,
  Alexander~C. Berg, and Li~Fei-Fei.
\newblock {ImageNet Large Scale Visual Recognition Challenge}.
\newblock {\em IJCV}, 115(3):211--252, 2015.

\bibitem{he2016deep}
Kaiming He, Xiangyu Zhang, Shaoqing Ren, and Jian Sun.
\newblock Deep residual learning for image recognition.
\newblock In {\em CVPR}, 2016.

\bibitem{loshchilov2018fixing}
Ilya Loshchilov and Frank Hutter.
\newblock Fixing weight decay regularization in adam.
\newblock In {\em ICLR}, 2018.

\end{thebibliography}
}

\end{document}